%% file: main.tex
\documentclass[10pt,twocolumn,letterpaper]{article}

\usepackage[pagenumbers]{cvpr} 

\definecolor{cvprblue}{rgb}{0.21,0.49,0.74}
\usepackage[pagebackref,breaklinks,colorlinks,allcolors=cvprblue]{hyperref}
\usepackage{colortbl}
\usepackage{multirow}
\usepackage{amssymb}
\usepackage{amsfonts}
\usepackage{bbm}
\usepackage{makecell}
\usepackage{graphicx}
\usepackage{float}
\usepackage{tcolorbox}
\tcbuselibrary{breakable}
\usepackage{etoolbox}
\usepackage{xr}
\hypersetup{
    colorlinks=true,
    urlcolor=magenta
}

\def\paperID{} 
\def\confName{CVPR}
\def\confYear{2026}

\title{MM-ReCoder: Advancing Chart-to-Code Generation with Reinforcement Learning and Self-Correction}

\author{Zitian Tang$^{1}\thanks{Work completed during internship at Amazon.}$ \quad Xu Zhang$^2$\quad Jianbo Yuan$^2$\quad Yang Zou$^2$\\
Varad Gunjal$^2$\quad Songyao Jiang$^2$\quad Davide Modolo$^2$\\
$^1$Brown University \quad$^2$Amazon AGI
\vspace{1mm}\\
\url{https://zitiantang.github.io/MM-ReCoder}
}

\makeatletter
\g@addto@macro\@maketitle{
\vspace{-35pt}
\begin{figure}[H]
\setlength{\linewidth}{\textwidth}
\setlength{\hsize}{\textwidth}
\centering
\includegraphics[trim={0cm, 0cm, 0cm, 0.0cm},clip,width=0.9\linewidth]{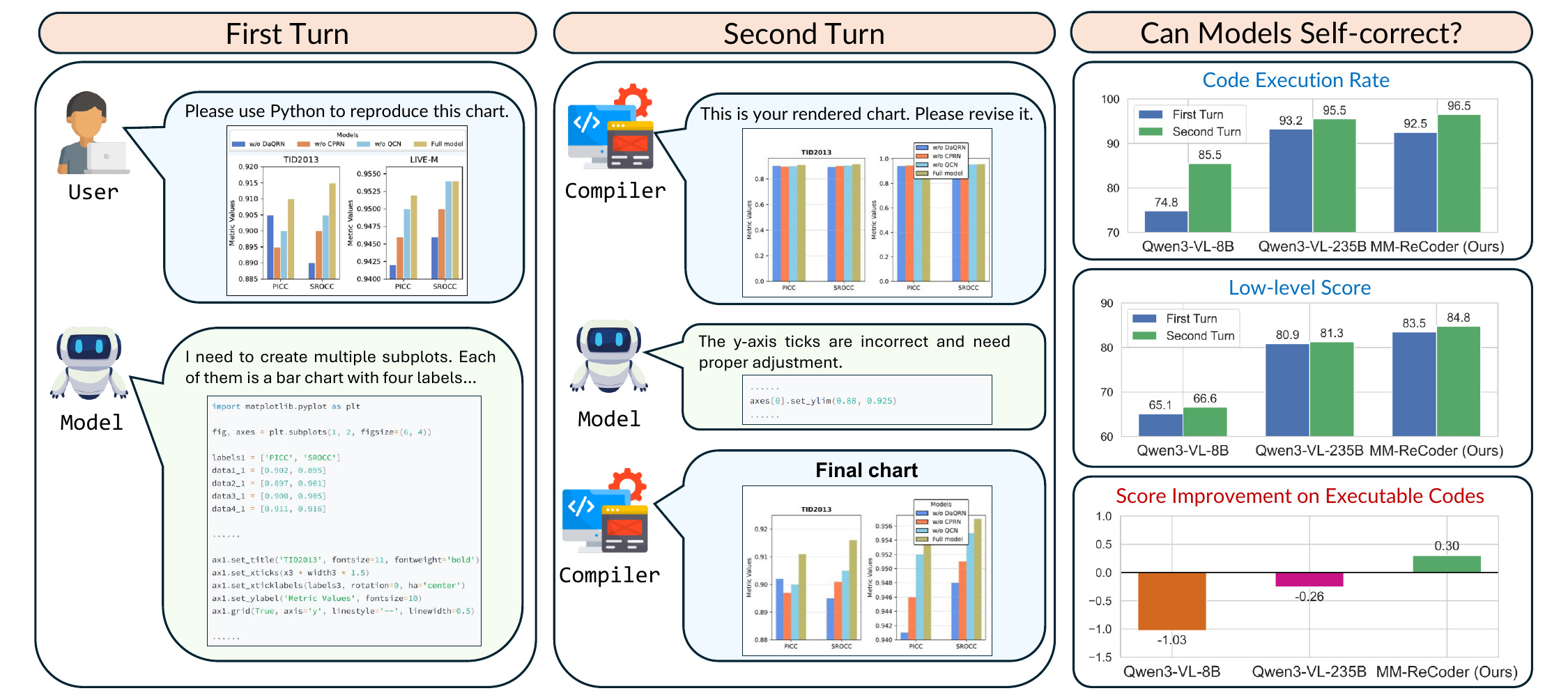}
\vspace{-10pt}
\caption{
{In multi-turn Chart2Code, the model takes the execution result from its first-turn code and revises the code accordingly. Although existing MLLMs improve evaluation scores between the two turns, the gains mainly come from increased code executability. When we restrict the analysis to cases where the first-turn code is already executable, existing models show a negative improvement, whereas our model demonstrates a positive one.
}
\vspace{-1mm}
}
\label{fig:teaser}
\end{figure}
}
\makeatother

\begin{document}
\maketitle
\input{sec/0_abstract}    
\input{sec/1_intro}
\input{sec/2_related_works}
\input{sec/3_method}

\input{sec/4_experiments}
\input{sec/5_conclusions}

{
    \small
    \bibliographystyle{ieeenat_fullname}
    \bibliography{main}
}

\input{sec/X_suppl}

\end{document}

%% file: sec/0_abstract.tex
\begin{abstract}
Multimodal Large Language Models (MLLMs) have recently demonstrated promising capabilities in multimodal coding tasks such as chart-to-code generation. However, existing methods primarily rely on supervised fine-tuning (SFT), which requires the model to learn code patterns through chart-code pairs but does not expose the model to a code execution environment. Moreover, while self-correction through execution feedback offers a potential route to improve coding quality, even state-of-the-art MLLMs have been shown to struggle with effective self-correction. In this work, we introduce MM-ReCoder, a chart-to-code generation model trained with reinforcement learning (RL) and equipped with self-correction ability. We propose a two-stage multi-turn self-correction RL strategy based on Group Relative Policy Optimization (GRPO). The first stage enhances the model's self-correction ability via rolling out a shared first turn, while the second stage improves the coding capability with full-trajectory optimization. MM-ReCoder learns to produce more accurate and executable code through the interaction with the environment and by iteratively correcting its own outputs. Our results on three chart-to-code benchmarks demonstrate the state-of-the-art performance of MM-ReCoder.
\end{abstract}

%% file: sec/1_intro.tex
\section{Introduction}
\label{sec:intro}

Scientific charts play a crucial role in helping humans interpret complex information by highlighting trends, relationships, and comparisons. 
However, manually creating these charts through graphical user interfaces or by writing code is a tedious and time-consuming process. To address these traditional limitations in chart generation and unlock powerful new applications, Multimodal LLMs have recently been explored and trained for automatic generation of chart code (Chart2Code task), which allows for the automation of the visualization process. In detail, in this new task, a chart is provided as input in the form of an image, and the model is expected to generate executable code to reproduce the chart. 

Existing Chart2Code approaches ~\cite{han2023chartllama,xia2025chartxchartvlm,zhang2024tinychart,meng2024chartassi,zhao2025chartcoder,si2024design2code,rodriguez2024starvector} typically collect a large corpus of image-code pairs and train the MLLMs via supervised fine-tuning (SFT).
While these methods demonstrate promising results, next token prediction alone doesn't guarantee that the generated code has executable syntax and visual details that are faithful to the references~\cite{tan2025chartmaster}.
To improve the execution success rate and the visual quality of the rendered content, recent works have begun exploring reinforcement learning (RL)~\cite{tan2025chartmaster,zhao2025vincicoder}. For instance, ChartMaster~\cite{tan2025chartmaster} introduced a combination of low-level reward (targeting chart elements such as chart type, color, and legends) and high-level rewards (encouraging visual similarity between the input and rendered images), yielding significant gains over SFT baselines.
One limitation of these RL-based Chart2Code methods is that they treat the problem as a one-shot generation task, meaning that they train their models to produce the final code in a single pass, without executing or refining it based on any feedback from the rendered result. 
Humans, conversely, operate iteratively: they implement code, execute it, visualize the results, and refine the code until the output meets expectations. This iterative {\it self-correcting} process is critical to reach the correct outcome, and in this paper we introduce an RL phase modeled after this process.
However, recent studies have shown that existing open-source MLLMs struggle to self-correct on multimodal coding tasks, even when provided with explicit feedback~\cite{yang2024chartmimic,si2024design2code}. 

To mitigate these limitations and help MLLMs successfully improve themselves in an iterative process, we propose {\bf MM-ReCoder}. 
MM-ReCoder follows a two-stage training curriculum: 1) SFT cold-start with ground-truth chart-code pairs and carefully crafted multi-turn coding data and 2) multi-turn self-correction RL.
The cold-start stage provides the model with basic coding knowledge and the ability to generate multi-turn coding responses, but it is not sufficient to guarantee reliable improvements in the second turn. 
To empower the model with reliable self-correction abilities, we introduce a two-stage multi-turn GRPO (Group Relative Policy Optimization)~\cite{shao2024deepseekmath} framework that extends standard GRPO to support self-correction.
In the first stage of this RL phase, we fix an online sampled first-turn output for each group and roll out multiple second-turn candidates, allowing the model to explore diverse refinement strategies.
In the second stage we then roll out both first-turn and second-turn outputs for each output in the group to further enhance coding performance. 

MM-ReCoder outperforms existing similar size Chart2Code models trained with the same dataset. For example, in ChartMimic~\cite{yang2024chartmimic}, it outperforms ChartCoder~\cite{zhao2025chartcoder} by 9.1\% in low-level score and 10.9\% in high-level score. Remarkably, with only 7B parameters, MM-ReCoder achieves performance comparable to much larger models such as Qwen3-VL-235B-A22B~\cite{qwen3vl_blog}.

\noindent In summary, our contributions are three-fold:
\begin{itemize}
    \item We propose MM-ReCoder, the first multimodal coding MLLM equipped with self-correction abilities. Trained with a framework containing cold start and two-stage RL, MM-ReCoder achieves state-of-the-art performance on chart-to-code  benchmarks~\cite{yang2024chartmimic,wu2024plot2code,xia2025chartxchartvlm} with $86.5\%$ low-level score on ChartMimic and $63.2\%$ text-match score on Plot2Code.
    \item We explore different RL training strategies for multi-turn self-correction. We find that shared-first-turn optimization followed by full-trajectory optimization can elicit the model's self-correction capability while improving its overall coding ability.
    \item We explore rule-based and model-based rewards. While rule-based reward helps the model faithfully reproduce chart elements, model-based reward effectively improves the visual quality of the reproduced images.
\end{itemize}

%% file: sec/2_related_works.tex
\section{Related Works}
\label{sec:related_works}

\begin{figure*}[t]
    \centering
    \includegraphics[width=0.85\linewidth]{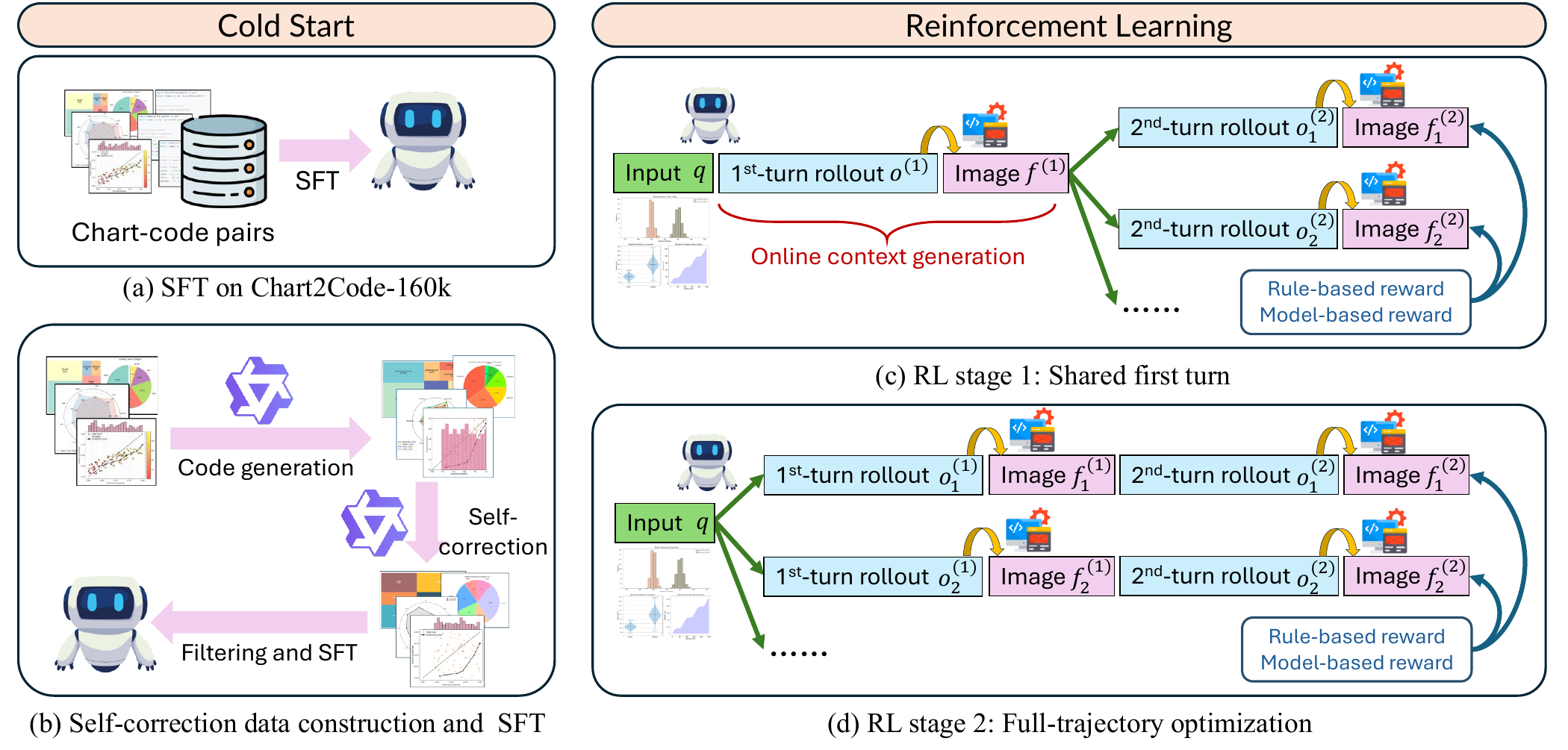}
    \vspace{-8pt}
    \caption{Training pipeline of MM-ReCoder. We first conduct two stages of cold start: (a) we first train the model on ground truth chart-code pairs with SFT, then (b) we construct self-correction data with Qwen3VL-235B~\cite{qwen3vl_blog}, filter the successful ones, and train our model on the filtered data. After cold start, we conduct two stages of reinforcement learning: (c) we first enhance the model's self-correction capability in the second turn via shared-first-turn optimization, then (d) we optimize the two turns jointly to improve the coding ability.}
    \label{fig:pipeline}
    \vspace{-15pt}
\end{figure*}

\subsection{Multimodal Coding}

In multimodal coding, models generate codes according to given multimodal inputs.
For example, in graphics derendering tasks, a model generates Python/HTML code to reproduce a given chart/webpage screenshot.
Recently, many benchmarks and datasets have been built for chart-to-code generation \cite{yang2024chartmimic,wu2024plot2code,xia2025chartxchartvlm,kondic2025chartgenscalingchartunderstanding,zhao2025chartcoder}, webpage-to-code generation~\cite{si2024design2code,yun2024web2code,laurencon2024unlocking}, and SVG-to-code generation~\cite{rodriguez2024starvector}.

With the rise of MLLMs, many works have trained domain-specific MLLMs tackling multiple tasks, \eg, chart description, QA, and redrawing~\cite{han2023chartllama,xia2025chartxchartvlm,zhang2024tinychart,meng2024chartassi}. Most of existing works~\cite{han2023chartllama,xia2025chartxchartvlm,zhang2024tinychart,meng2024chartassi,zhao2025chartcoder,si2024design2code,rodriguez2024starvector} consider Chart2Code problem as SFT problem. For example, ChartCoder~\cite{zhao2025chartcoder} collects 160k chart-code pairs and uses them to train an MLLM.
Beyond SFT, \citet{zhang2025boostingcharttocode} proposes dual preference-guided refinement to iteratively collect code preference pairs and refine a model with offline reinforcement learning. ChartMaster~\cite{tan2025chartmaster} first proposes to use GRPO with low-level and high-level rewards to improve chart quality. VinciCoder~\cite{zhao2025vincicoder} develops a single model to address all Vision2Code problems via SFT and RL. However, all existing RL-based approaches treat Chart2Code as a one-shot generation task, preventing models from iteratively refining their code to improve generation quality. 

\subsection{Reinforcement Learning for MLLMs}

Reinforcement learning (RL) has recently emerged as an effective paradigm for improving reasoning and alignment capabilities in LLMs.
Early methods such as REINFORCE~\cite{ahmadian2024basicsrevisitingreinforcestyle}, PPO~\cite{schulman2017ppo,ouyang2022traininglanguagemodelsfollow}, and DPO~\cite{rafailov2024dpo} established the foundation for optimizing LLMs through reward-guided policy gradients.
DeepSeek-R1~\cite{deepseekai2025deepseekr1} first demonstrated that RL can substantially improve the reasoning ability of LLMs via GRPO~\cite{shao2024deepseekmath} with verifiable rewards.
R1-style thinking patterns, \eg, self-verification, emerge during RL training.

Inspired by these advances in LLMs, recent works extend RL techniques to MLLMs to enhance multimodal reasoning~\cite{huang2025visionr1,yang2025r1onevision,wei2025openvisionreasoner,shen2025vlm}.
They find cold start essential to reproduce the R1-style thinking patterns on MLLMs.
Vision-R1~\cite{huang2025visionr1} and R1-Onevision~\cite{yang2025r1onevision} generates multimodal R1-style cold start data by applying DeepSeek-R1 on dense image captions, while OVR~\cite{wei2025openvisionreasoner} shows that R1-style reasoning learned from linguistic cold start can be transferred to multimodal tasks.
In contrast, VL-Rethinker~\cite{vl-rethinker} appends a rethinking trigger token after the model output to force self-reflection during RL.
Moreover, some other works extend RL to video reasoning~\cite{feng2025video,zhang2025tinyllava,li2025videochatr1,yan2025videochatr15} or to train MLLMs for tool calling~\cite{zheng2025deepeyes}.
These efforts collectively indicate the effectiveness of RL in improving reasoning in MLLMs.

\subsection{Self-correction for LLMs and MLLMs}

Self-correction aims to prompt the models to critique and refine its own answers.
While training-free prompt-based self-correction is shown to be ineffective for LLMs~\cite{huang2024largelanguagemodelsselfcorrect,qu2024recursiveintrospectionteachinglanguage,tyen2024llmsreasoningerrorscorrect,zheng2024naturalplanbenchmarkingllms}, one way to enable self-correction is SFT on annotated or generated self-correction data~\cite{saunders2022selfcritiquingmodelsassistinghuman,qu2024recursiveintrospectionteachinglanguage}.
However, \citet{kumar2024traininglanguagemodelsselfcorrect} show that self-correction learned by SFT suffers from distribution shift and behavior collapse. They propose SCoRe to elicit the self-correction ability of LLMs with two-stage RL.
Moreover, a few works explore self-correction in coding tasks~\cite{cho2025selfcorrectingcodegenerationusing,gehring2025rlefgroundingcodellms,jain2025multiturn}.
For example, CoCoS~\cite{cho2025selfcorrectingcodegenerationusing} elicits self-correction by carefully setting the reward discount factor of the first turn.
While these methods use REINFORCE~\cite{ahmadian2024basicsrevisitingreinforcestyle} and PPO~\cite{schulman2017ppo,ouyang2022traininglanguagemodelsfollow} as the RL algorithm, we explore training self-correction with GRPO.

In multimodal coding, ChartMimic~\cite{yang2024chartmimic} and Design2Code~\cite{si2024design2code} demonstrate that only very large MLLMs are able to self-correct their own solutions.
We are the first to explore eliciting MLLM self-correction in this domain.

%% file: sec/3_method.tex
\section{Method}
\label{sec:method}

\subsection{Problem Formulation}

In Chart2Code task, given a chart image $q$, a model is asked to output $o$, which includes a thinking trace and a Python code, to reproduce input image $q$. In conventional single-shot generation~\cite{han2023chartllama,xia2025chartxchartvlm,zhang2024tinychart,meng2024chartassi,zhao2025chartcoder}, the output $o$ is directly sampled from the model output distribution $\pi_\theta(O|q)$, where $\theta$ is the model parameter and $O$ is the output space.

For self-correction, without loss of generality, we only consider one correction turn. In addition to directly sampling the first turn solution $o^{(1)}$ from the model distribution. We execute the code in $o^{(1)}$ to render an output image $I^{(1)}$.
The output image will be sent back to MLLM for self-correction. 
Specifically, we construct the feedback $f^{(1)}$ as \textit{``$I^{(1)}$ Here is the image rendered by your code. Please continue modifying the code.''}. If the code is not executable, $f^{(1)}$ becomes \textit{``\{error information\} Your code encountered a runtime error. Please continue modifying the code.''} (See the complete prompts in the supplementary).
Then the model samples a second-turn output conditioned on the first turn and the feedback $o^{(2)}\sim\pi_\theta(O|q,o^{(1)},f^{(1)})$. This process can be iteratively performed for multiple rounds. 

\subsection{Self-Correction Ability of Off-the-Shelf Models}
\label{sec:self_correction_off_the_shelf}

We start by testing the self-correction ability of the most recent off-the-shelf MLLMs.
We evaluate Qwen3-VL-8B and Qwen3-VL-235B-A22B~\cite{qwen3vl_blog} on ChartMimic~\cite{yang2024chartmimic}
in~\Cref{fig:teaser} (more details in \Cref{tab:self_correction_detailed_metrics}).
Qwen3-VL-8B improves the overall low-level score by $1.5\%$ between two turns while Qwen3-VL-235B improves it by $0.4\%$.
However, we notice that the code execution rates increase more significantly ($10.7\%$ and $2.3\%$).
This suggests that the observed gains in low-level scores may come from executing more charts rather than refining already-executable code.

To validate this hypothesis, we filter the images that are successfully rendered in both turns and compute their average score improvement.
As \Cref{fig:teaser} shows, their improvements on executable codes are $-1.03\%$ and $-0.26\%$, respectively, indicating that the models are unable to refine charts that were already executable in the first turn. Inspired by this, we propose MM-ReCoder which focuses on iteratively improving the visual quality of the generated image.

\subsection{MM-ReCoder}

To achieve the self-correction capability, the proposed MM-ReCoder training pipeline has two phases: cold start and multi-turn self-correction RL, illustrated in~\Cref{fig:pipeline}.

In the cold start phase, we first train the model with SFT on Chart2Code-160k~\cite{zhao2025chartcoder}, which includes 160k chart-code pairs.
While this effectively enhances the model's coding capability, the model loses its multi-turn ability after training on single-turn data (as more than 80\% of the second turns are repeating the first turns, see \Cref{tab:self_correction_of_cold_start}).
Therefore, in the second cold start stage, we construct and train our model on 7k two-turn self-correction data using Qwen3-VL-235B-A22B-Instruct~\cite{qwen3vl_blog} on the images from Chart2Code-160k.
Specifically, we first generate a two-turn conversation on each image. Then we run rejection sampling and only keep samples whose second-turn low-level score exceeds the first turn by at least 0.02, a threshold we find sufficient to visually observe chart improvements.

Although this cold-start stage enables the MLLM to produce multi-turn coding responses, it does not reliably ensure improvement in the second turn. To address this, we propose a two-stage multi-turn self-correction RL using GRPO.
We first perform shared-first-turn optimization to improve the model's self-correction capability, then conduct full-trajectory optimization to boost its overall coding performance.
We introduce the details in the following.

\subsection{Multi-Turn Self-Correction RL}

We start by revisiting single-turn on-policy GRPO.
For each query $q$, GRPO samples a group of outputs $\{o_1, o_2, \cdots,o_G\}$ following the policy $\pi_{\theta}$.
The optimization objective can be derived as\footnotemark{}
\begin{equation}
\small
\begin{aligned}
\mathcal{J}_{GRPO}(\theta) &= \mathbb{E}
\Biggl[
\frac1G \sum_{i=1}^G
    \frac{\pi_{\theta}(o_i | q)}{\mathrm{SG}[\pi_{\theta}(o_i | q)]}A_i
\Biggr],
\end{aligned}
\label{eq:grpo_ours}
\end{equation}
where $\mathrm{SG}[\cdot]$ denotes stop gradient and $A_i$ is the advantage of $o_i$. $A_i$ can be derived as 
\begin{equation}
\small
A_i=\frac{R(o_i)-\mathrm{mean}(\{R(o_1),R(o_2),\cdots,R(o_G)\})}{\mathrm{std}(\{R(o_1),R(o_2),\cdots,R(o_G)\})},
\label{eq:advantage}
\end{equation}
while $R(\cdot)$ is a reward function taking MLLM output as input to produce a scalar reward. For multi-turn self-correction, we first draw the first turn $o_i^{(1)}\sim\pi_\theta(O|q)$ followed by the second-turn output $o_i^{(2)}\sim\pi_\theta(O|q,o_i^{(1)},f_i^{(1)})$.
Overall, our RL formulation has two key components - the reward function design and the multi-turn roll-out strategy.
\footnotetext{To simplify the notation, we omit the token-level averaging in GRPO throughout the paper.}

\subsubsection{Reward Design}
\label{sec:reward_design}

\begin{figure}[t]
    \centering
    \includegraphics[width=0.95\linewidth]{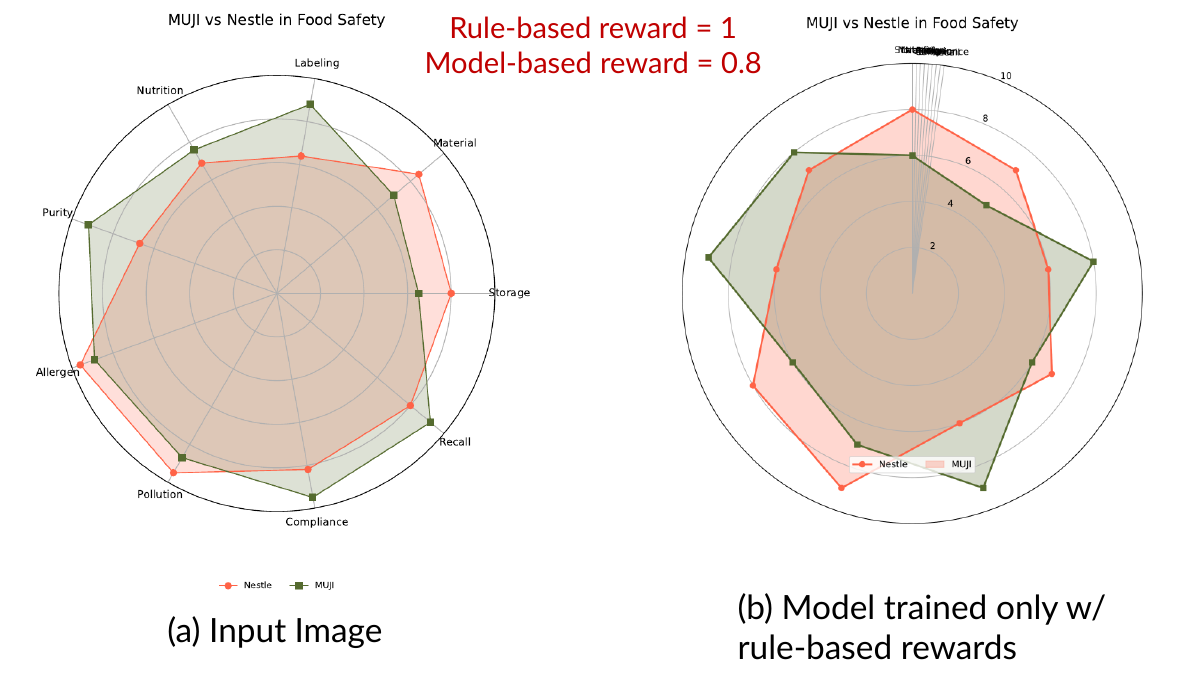}
    \vspace{-10pt}
    \caption{Result of a model trained solely with rule-based reward. The model receives a full rule-based reward though the texts are overlapped. But the model-based reward can penalize this chart.}
    \label{fig:reward_hacking}
    \vspace{-15pt}
\end{figure}

Designing rewards for Chart2Code task is inherently challenging since there is no perfect metric that fully captures visual similarity and semantic correctness between two charts.
In common practices, one may extract the visual elements (\eg, text and color) from the rendered image and the reference image to calculate rule-based rewards, or use VLMs and MLLMs to evaluate the image similarity~\cite{liu2025flow}.
In this work, we leverage both perspectives.
For the Chart2Code task, ChartMimic~\cite{yang2024chartmimic} introduces rule-based low-level metrics and a model-based high-level metric for evaluation purpose. We reuse them as our RL rewards.

\noindent\textbf{Rule-based reward.}
Given the python code for rending a chart, we can hook into the Matplotlib functions and execute the code to obtain the visual elements.
For example, to extract all the texts rendered by the code, we wrap the \texttt{RendererPdf.draw\_text} function with a custom function that intercepts every text-drawing call, records the text content, and forwards the call to the original renderer.
By this approach, we extract the chart type, layout, text, and color from the generated code and ground truth code and measure their similarity.
Specifically, we calculate the F1-score for chart type and text, CIE Lab color distance for color, and compare the number of rows and columns for layout.
While each metric is in range $\left[0, 1\right]$, we take their average as the overall rule-based reward.

Although the rule-based reward can provide a precise score from visual elements, it fails to measure many aspects of chart quality.
If a model is trained solely with rule-based rewards, it may suffer from issues such as overlapping elements, inaccurate data points, or improperly sized fonts.
For example, in \Cref{fig:reward_hacking}, a chart containing overlapping text can still receive a rule-based reward of 1.

\noindent\textbf{Model-based reward.}
In contrast to rule-based rewards, the model-based reward measures both visual and semantic similarity. While ChartMimic~\cite{yang2024chartmimic} calls GPT-4o to score the similarity between two charts, it is computationally expensive for large-scale RL.
Instead, we employ an open-source MLLM, Qwen2.5-VL-72B~\cite{bai2025qwen25vltechnicalreport}, as the reward model.
Given a reproduced image and the reference image, we ask the model to evaluate the reproduced image from six aspects: chart type, layout, text, data, style, and clarity.
Each has a maximum score of either 10 or 20 points, leading to a total score of 100.
The total score is scaled to the range of $\left[0, 1\right]$ and used as the model-based reward.

Finally, to encourage the model to think before coding, we include a \textbf{Format reward}, which is 1 if the model output follows \texttt{<think>...</think>```python...```}, otherwise 0.
We combine the rewards by \texttt{$(1-\alpha-\beta)$* Format + $\alpha$*Rule-based + $\beta$*Model-based}, where $\alpha$ and $\beta$ are hyperparameters.
Ablation studies on the rewards weights are in~\Cref{sec:reward_weight}.

\subsubsection{Roll-out Strategies for Self-correction}
\label{sec:training_strategy}
While self-correction can be performed for multiple rounds, we focus on training two-turn self-correction during RL, \emph{i.e.}, we only optimize the model's first-turn output $o^{(1)}$ and its refinement $o^{(2)}$.
At inference time, however, the model can perform an arbitrary number of self-correction turns.
To train a multi-turn model, we consider two roll-out strategies:

\noindent\textbf{Full-Trajectory Optimization.}
As shown in~\Cref{fig:pipeline}(d), we jointly optimize both the first and second turns in each trajectory. For each input image, we independently roll out $G$ trajectories with both the first- and second-turn generations.
Specifically, we first sample $o_i^{(1)}\sim\pi_\theta(O|q)$ and then $o_i^{(2)}\sim\pi_\theta(O|q,o_i^{(1)},f_i^{(1)})$.
The reward is computed over the entire trajectory, and the policy is updated to maximize
\begin{equation}
\small
\begin{aligned}
\mathcal{J}^{(full)}_{GRPO}(\theta) &= \mathbb{E}
\Biggl[
\frac1G \sum_{i=1}^G
    \frac{\pi_{\theta}(o_i^{(1)} | q)\pi_{\theta}(o_i^{(2)} | q,o_i^{(1)},f_i^{(1)})}{\mathrm{SG}[\pi_{\theta}(o_i^{(1)} | q)\pi_{\theta}(o_i^{(2)} | q,o_i^{(1)},f_i^{(1)})]}A_i
\Biggr].
\end{aligned}
\label{eq:full_traj_opt}
\end{equation}

As there are two turns, besides the final turn, we can also use the first turn in reward calculation. Additionally, we can further include a boosting reward to encourage the model to generate improved second turn compared to the first turn.
Specifically, the reward for a trajectory $(o_i^{(1)},o_i^{(2)})$ is
\begin{equation}
\small
R(o_i^{(1)}, o_i^{(2)}) =  R(o_i^{(2)}) + \gamma\cdot R(o_i^{(1)}) + \eta\cdot\textrm{B}(R(o_i^{(1)}), R(o_i^{(2)})),
\label{eq:full_reward}
\end{equation}
where $\gamma$ is a hyperparameter to balance the rewards of the two turns, $\eta$ is a hyperparameter to weight the boosting reward $\textrm{B}(R(o_i^{(1)}),R(o_i^{(2)}))=\mathbbm{1}(R(o_i^{(2)}) > R(o_i^{(1)}))$. $\mathbbm{1}(\cdot)$ is indicator function to encourage the second-turn reward to be higher than the first-turn reward.

\noindent\textbf{Shared First Turn.}
As \Cref{fig:pipeline}(c) shows, for each input, we first roll out a shared first-turn output $o^{(1)}\sim\pi_\theta(O|q)$ with feedback $f^{(1)}$.
We then generate multiple second-turn refinements $o_i^{(2)}\sim\pi_\theta(O|q,o^{(1)},f^{(1)})$ conditioned on this shared first turn. 
The optimization objective is
\begin{equation}
\small
\hspace*{-5.7pt}
\begin{aligned}
\mathcal{J}^{(shared)}_{GRPO}(\theta) &= \mathbb{E}
\Biggl[
\frac1G \sum_{i=1}^G
    \frac{\pi_{\theta}(o_i^{(2)} | q,o^{(1)},f^{(1)})}{\mathrm{SG}[{\pi_{\theta}(o_i^{(2)} | q,o^{(1)},f^{(1)})]}}A_i
\Biggr].
\end{aligned}
\label{eq:turn_wise_opt}
\end{equation} 
Note that we treat the first turn as shared context. As a result, RL optimization is applied only to the second turn. Interestingly, we observe that the model’s first-turn capability also improves during this training process. More importantly, the first-turn context is sampled online during training, which gradually increases the difficulty of generating the second-turn outputs as the model improves.

\noindent\textbf{Two-Stage Training.}
In practice, we find that combining the shared–first-turn strategy with the full-trajectory strategy yields the best performance. Specifically, we begin with the shared first-turn roll-out, which encourages the model to develop strong self-correction capabilities by focusing on diverse second-turn refinements. To further improve overall performance, we introduce the second stage that uses the full-trajectory roll-out, allowing the model to jointly optimize both first-turn and second-turn outputs. For reward computation, we observe that calculating the reward only at the final turn (\emph{i.e.}, $\gamma=0$) leads to the best performance.

%% file: sec/4_experiments.tex
\section{Experiments}
\label{sec:experiments}

\begin{table*}[t]
\small
\centering
\resizebox{\textwidth}{!}{
\begin{tabular}{c|c|ccc|ccc|c}
\toprule
\multirow{3}{*}{Model} & \multirow{3}{*}{\# Turns} & \multicolumn{3}{c|}{ChartMimic} & \multicolumn{3}{c|}{Plot2Code} & ChartX \\
\cmidrule{3-9} 
  & & \multirow{2}{*}{Exec.Rate} & \multirow{2}{*}{Low-Level} & High-Level & \multirow{2}{*}{Pass Rate} & \multirow{2}{*}{Text-Match}  & Rating  & GPT-score  \\ 
  & & & & {\footnotesize\textcolor{gray}{(GPT-4o)}} & & & {\footnotesize\textcolor{gray}{(GPT-4V/4o)}} & {\footnotesize\textcolor{gray}{(GPT-4/4o)}} \\
\midrule
Full score & - & 100 & 100 & 100 & 100 & 100 &10 & 5 \\
\midrule
\rowcolor[gray]{0.9} \multicolumn{9}{c}{\it{Proprietary}} \\ 
\midrule
GeminiProVision & 1 & 68.2 & 53.8 & 53.3  & 68.2 & 53.6& 3.69 / -\;\;\;\;\; &-\\
Claude-3-opus & 1 & 83.3 & 60.5 & 60.1 & 84.1 & 57.5 & 3.80 / -\;\;\;\;\; & -\\
GPT-4V & 1 & 91.2 & 76.4 & 78.9 & 84.1 & 57.7 & 5.58 / -\;\;\;\;\; & 2.63 / -\;\;\;\;\; \\
GPT-4o & 1 & 93.2 & 79.0 & 83.5 & 88.6 & 56.3 & 5.71 / -\;\;\;\;\; &-\\
\rowcolor[rgb] {1, 1, 0.848} GPT-4o (reproduced) & 1 & 94.0 & 81.8 & 83.7 & 82.6 & 59.8 & \;\;\;\;\;- / 5.42 & \;\;\;\;\;- / 2.08 \\
\rowcolor[rgb] {1, 1, 0.848} GPT-4o (reproduced) & 2 & 96.7 & 82.9 & 86.8 & 90.2 & 58.4 & \;\;\;\;\;- / \textbf{6.12} & \;\;\;\;\;- / 2.07 \\
\midrule
\rowcolor[gray]{0.9} \multicolumn{9}{c}{\it{Open-Source General-Domain}} \\
\midrule
DeepSeek-VL-7B~\cite{lu2024deepseekvl} & 1 & 41.3 & 19.0 & 17.6 & 64.4 & 32.6& 2.26 / -\;\;\;\;\; & -\\
LLaVA-Next-Mistral-7B~\cite{li2024llavanext} & 1 & 59.7 & 20.7 & 21.3 & 72.0 & 38.7 & 2.87 / -\;\;\;\;\;  &- \\
MiniCPM-Llama3-V2.5~\cite{yao2024minicpmv} & 1 & 80.3 & 36.6 & 42.1  & 76.3 & 37.3& 2.61 / -\;\;\;\;\; &1.66 / -\;\;\;\;\; \\
InternVL2-8B~\cite{chen2024far} & 1 & 61.8 & 34.4 & 38.9 & 77.3 & 37.1 & 2.78 / -\;\;\;\;\; & 1.63 / -\;\;\;\;\; \\
InternVL2-26B~\cite{chen2024far} & 1 & 69.3 & 41.4 & 47.4  & 81.3 & 43.1 & 3.42 / -\;\;\;\;\; & 1.70 / -\;\;\;\;\; \\
InternVL2-Llama3-76B~\cite{chen2024far} & 1 & {83.2} & {54.8} & {62.2}  & {85.6} & 46.6 & 3.89 / -\;\;\;\;\; & 1.74 / -\;\;\;\;\; \\
Qwen2-VL-7B~\cite{Qwen2VL} & 1 & 67.0 & 32.9 & 35.0 & 68.2 & 33.8 & 3.10 / -\;\;\;\;\; & 1.50 / -\;\;\;\;\; \\
Qwen2-VL-72B~\cite{Qwen2VL} & 1 & 73.3 &54.4  & 50.9 & 72.0 & {53.4}  & {4.26} / -\;\;\;\;\;  & 1.69 / -\;\;\;\;\; \\
\rowcolor[rgb] {0.9, 1, 1} Qwen2.5-VL-7B ~\cite{bai2025qwen25vltechnicalreport} & 1 & 71.5 & 56.2 & 49.6 & 75.0 & 47.8 & \;\;\;\;\;- / 3.51 & \;\;\;\;\;- / 1.99 \\
\rowcolor[rgb] {0.9, 1, 1} Qwen2.5-VL-7B ~\cite{bai2025qwen25vltechnicalreport} & 2 & 74.3 & 57.2 & 50.5 & 78.5 & 43.8 & \;\;\;\;\;- / 3.25 & \;\;\;\;\;- / 1.96 \\
\rowcolor[rgb] {1, 1, 0.848} Qwen2.5-VL-72B ~\cite{bai2025qwen25vltechnicalreport} & 1 & 89.5 & 72.3 & 74.7 & 82.6 & 55.0 & \;\;\;\;\;- / 4.67 & \;\;\;\;\;- / 2.28 \\
\rowcolor[rgb] {1, 1, 0.848} Qwen2.5-VL-72B ~\cite{bai2025qwen25vltechnicalreport} & 2 & 93.2 & 75.9 & 77.6 & 87.1 & 55.4 & \;\;\;\;\;- / 4.96 & \;\;\;\;\;- / 2.26 \\
\rowcolor[rgb] {0.9, 1, 1} Qwen3-VL-8B~\cite{qwen3vl_blog} & 1 & 74.8 & 65.1 & 66.2 & 73.5 & 49.0 & \;\;\;\;\;- / 4.11 & \;\;\;\;\;- / 2.31 \\
\rowcolor[rgb] {0.9, 1, 1} Qwen3-VL-8B~\cite{qwen3vl_blog} & 2 & 85.5 & 66.6 & 71.1 & 84.1 & 46.8 & \;\;\;\;\;- / 4.31 & \;\;\;\;\;- / 2.15 \\
\rowcolor[rgb] {1, 1, 0.848} Qwen3-VL-235B-A22B~\cite{qwen3vl_blog} & 1 & 93.2 & 80.9 & 85.9 & 90.2 & 60.9 & \;\;\;\;\;- / 5.93 & \;\;\;\;\;- / \textbf{2.65} \\
\rowcolor[rgb] {1, 1, 0.848} Qwen3-VL-235B-A22B~\cite{qwen3vl_blog} & 2 & 95.5 & 81.3 & \textbf{87.2} & 91.7 & 60.4 & \;\;\;\;\;- / 5.79 & \;\;\;\;\;- / 2.59\\
\midrule
\rowcolor[gray]{0.9} \multicolumn{9}{c}{\it{Open-Source Chart-Domain}} \\
\midrule
ChartLlama~\cite{han2023chartllama} & 1 & 57.5 & 24.8 & 28.1 & 58.4 & 40.3 & 2.32 / -\;\;\;\;\; & 0.94 / -\;\;\;\;\; \\
TinyChart~\cite{zhang2024tinychart} & 1 & 42.5 & 26.3 & 25.9 & 43.2 & 44.6 & 2.19 / -\;\;\;\;\; & {1.89} / -\;\;\;\;\; \\
ChartVLM-L~\cite{xia2025chartxchartvlm} & 1 & 19.5 & 15.8 & 13.9 & - & - & - &  1.58 / 0.99 \\
ChartCoder~\cite{zhao2025chartcoder} & 1 & {91.4} & {77.4} & {74.0}
 & {87.9} & 54.5 & 4.50 / 3.45 & {2.09} / 2.08 \\
MM-ReCoder-Single (Ours) & 1 & 95.0 & 84.3 & 83.7 & 89.4 & 62.8 &\;\;\;\;\;- / 5.10 & \;\;\;\;\;- / 2.24 \\
\rowcolor[rgb] {1, 0.9, 1}MM-ReCoder (Ours) & 1 & 92.5 & 83.5 & 81.2 & 88.6 & \textbf{63.2} & \;\;\;\;\;- / 4.62 & \;\;\;\;\;- / 2.31 \\
\rowcolor[rgb] {1, 0.9, 1}MM-ReCoder (Ours) & 2 & 96.5 & 84.8 & 84.7 & 97.7 & 62.2 & \;\;\;\;\;- / 5.17 & \;\;\;\;\;- / 2.32 \\
\rowcolor[rgb] {1, 0.9, 1} MM-ReCoder (Ours) & 4 & \textbf{97.5} & \textbf{86.5} & 84.9 & \textbf{98.5} & 62.7 & \;\;\;\;\;- / 5.24 & \;\;\;\;\;- / 2.31 \\
\bottomrule 
\end{tabular}}
\vspace{-5pt}
\caption{Evaluation results on the Chart2Code task. MM-ReCoder achieves the best low-level score on ChartMimic and Text-match score on Plot2Code. On the other evaluation metrics, MM-ReCoder outperforms models of comparable size significantly.}
\label{tab:main_results_chart}
\vspace{-15pt}
\end{table*}

\subsection{Implementation}

We use Qwen2.5-VL-7B~\cite{bai2025qwen25vltechnicalreport} as the base model of MM-ReCoder.
We train it on Chart2Code-160k~\cite{zhao2025chartcoder} in both the cold start and the reinforcement learning phases. 
For cold start, we train the model for one epoch on Chart2Code-160k and two epochs on the constructed self-correction data. 
We use the LLaMA-Factory framework~\cite{zheng2024llamafactoryunifiedefficientfinetuning} with a batch size of 128 and learning rate of $10^{-5}$ on 8 NVIDIA H200s.

For reinforcement learning, we use the VeRL framework~\cite{sheng2024hybridflow}.
We use GRPO with a group size of $G=8$ to train our model for one epoch in each RL stage.
The batch size is 128 and the learning rate is $10^{-6}$.
The maximum response length for each turn is 4,096 tokens.
Training is conducted on $16\times8$ NVIDIA H200s with another $4\times8$ H200s serving Qwen2.5-VL-72B as reward model.
The two RL stages take 61 and 73 hours, respectively.

\subsection{Benchmarks and Baselines}
Following prior work \cite{zhao2025chartcoder}, we evaluate MM-ReCoder on ChartMimic~\cite{yang2024chartmimic}, Plot2Code~\cite{wu2024plot2code}, and ChartX~\cite{xia2025chartxchartvlm}.
We report the evaluation metrics defined by each benchmark.
For ChartMimic, we include execution rate, the low-level score, and the high-level score.
For Plot2Code, we report pass rate, text-matching score, and the GPT-4 rating.
For ChartX, we report the GPT-score.
Because the GPT-4 and GPT-4V checkpoints used in prior Plot2Code and ChartX evaluations are no longer available through the OpenAI API, we use GPT-4o instead and evaluate several existing models as updated baselines. Please refer to the corresponding benchmark papers for detailed metric definitions.

We evaluate MM-ReCoder, GPT-4o, Qwen2.5-VL, and Qwen3-VL under both single-turn and multi-turn self-correction settings.
Due to computational constraints, other baselines are evaluated only in the single-turn setting.
Moreover, we develop a single-turn RL baseline MM-ReCoder-Single.
It uses the same backbone and cold start as MM-ReCoder but RL is conducted in a single-turn setting.

\begin{table*}[t]
\small
\centering
\resizebox{\textwidth}{!}{
\begin{tabular}{c|cc|cc|cc|cccc}
\toprule
\multirow{2}{*}{Strategy} & \multirow{2}{*}{$\gamma$} & \multirow{2}{*}{$\eta$} & \multicolumn{2}{c|}{First turn} & \multicolumn{2}{c|}{Second turn} &  \multirow{2}{*}{\makecell{Avg. low-level improvement \\ on rendered charts}} & \multirow{2}{*}{\makecell{Percentage of \\ improved samples}} & \multirow{2}{*}{\makecell{Percentage of \\ degraded samples}} & \multirow{2}{*}{\makecell{Percentage of \\ repeated code}} \\
\cmidrule{4-7} 
  & & & Exec.Rate & Low-Level & Exec.Rate & Low-Level &  \\
\midrule
                & 0   & 0   & 91.8 & 81.8 & 95.2 & 83.9 & 0.21 & 3.4 & 2.6 & 46.9 \\
Full-trajectory & 0   & 0.1 & 64.5 & 66.7 & 96.0 & 84.3 & 10.11 & 87.5 & 2.3 & 0.3 \\
                & 0.5 & 0.1 & 96.5 & 85.0 & 97.5 & 85.5 & -0.03 & 2.2 & 2.2 & 55.6 \\
\midrule
Turn-wise & -   & -   & 94.5 & 82.7 & 96.8 & 83.4 & 0.05 & 2.7 & 1.6 & 63.0 \\
\midrule
Shared-first-turn & -  & -   & 91.7 & 79.8 & 95.8 & 82.6 & 0.72 & 14.4 & 8.4 & 16.8 \\
\midrule
Shared-first-turn & 0   & 0   & 94.3 & 83.7 & 97.7 & 86.0 & 0.55 & 12.1 & 8.2 & 21.6 \\
 + Full-trajectory & 0.5 & 0.1 & 96.5 & 83.7 & 96.3 & 83.5 & -0.03 & 1.8 & 2.7 & 81.1 \\
\bottomrule 
\end{tabular}}
\vspace{-5pt}
\caption{Comparison of RL strategies for self-correction. Our two-stage strategy enables self-correction capability while the others cannot.}
\label{tab:compare_strategies}
\vspace{-12pt}
\end{table*}

\begin{table}[t]
\setlength{\tabcolsep}{3pt}
\small
\centering
\resizebox{0.49\textwidth}{!}{
\begin{tabular}{c|c|c|c|c|c|c}
\toprule
\multirow{3}{*}{Model} & \multicolumn{3}{c|}{Low-level} & \multicolumn{3}{c}{High-level} \\
\cmidrule{2-7}
  & \makecell{Average \\ improve} & \makecell{Improved$\%$} & \makecell{Degraded$\%$} & \makecell{Average \\ improve} & \makecell{Improved$\%$ \\ \symbol{64}0.1} & \makecell{Degraded$\%$ \\ \symbol{64}0.1} \\
\midrule
Qwen2.5-VL-7B & -0.36 & 8.1 & 13.9 & -0.85 & 8.9 & 10.7 \\
Qwen2.5-VL-72B & 0.22 & 15.9 & 14.5 & 0.20 & 11.2 & 10.2 \\
Qwen3-VL-8B & -1.03 & 6.0 & 14.5 & -0.70 & 6.8 & 8.3 \\
Qwen3-VL-235B-A22B & -0.26 & 13.4 & 13.5 & 0.27 & 5.4 & 4.8 \\
GPT-4o & 0.85 & 22.4 & 12.0 & 1.76 & 11.2 & 4.9 \\
MM-ReCoder (Ours) & 0.30 & 7.3 & 4.5 & 0.89 & 7.1 & 4.3 \\
\bottomrule 
\end{tabular}}
\vspace{-5pt}
\caption{Self-correction capability of MM-ReCoder and baselines.}
\label{tab:self_correction_detailed_metrics}
\vspace{-5pt}
\end{table}

\begin{table}[t]
\setlength{\tabcolsep}{3pt}
\small
\centering
\resizebox{0.48\textwidth}{!}{
\begin{tabular}{c|ccc|cc|cc}
\toprule
\multirow{2}{*}{Turn} & \multirow{2}{*}{\makecell{Exec.\\Rate}} & \multirow{2}{*}{\makecell{Low-\\level}} & \multirow{2}{*}{\makecell{High-\\level}} & \multicolumn{2}{c|}{Low-level} & \multicolumn{2}{c}{High-level} \\
\cmidrule{5-8}
& & & & Avg. improve & Improved\% & Avg. improve & Improved\%\symbol{64}0.1 \\
\midrule
1 & 92.5 & 83.5 & 81.2 & - & - & - & - \\
2 & 96.5 & 84.8 & 84.7 & 0.30 & 7.3 & 0.89 & 7.1 \\
3 & 96.5 & 85.3 & 84.8 & 0.12 & 3.8 & 0.22 & 3.7 \\
4 & 97.5 & 86.5 & 84.9 & 0.10 & 2.6 & -0.1 & 2.6 \\
5 & 98.3 & 86.4 & 85.5 & -0.12 & 2.4 & 0.26 & 3.7 \\
\bottomrule 
\end{tabular}}
\vspace{-5pt}
\caption{MM-ReCoder iteratively self-corrects for multiple turns.}
\vspace{-15pt}
\label{tab:iterative_self_correct}
\end{table}

\subsection{Main Results}

\Cref{tab:main_results_chart} shows the results of MM-ReCoder and baselines on Chart2Code tasks. 
MM-ReCoder not only significantly outperforms chart-domain models and models of comparable size, but also surpasses GPT-4o and Qwen3-VL-235B-A22B on ChartMimic low-level score and Plot2Code text-match score.
Compared to our base model Qwen2.5-VL-7B, even with only one turn, on ChartMimic, MM-ReCoder improves the execution rate by 29\%, the low-level score by $48\%$, and the high-level score by $63\%$.

Adding self-reflection further improves the model performance. By adding one correction turn, on ChartMimic, the execution rate can improve from 95\% to 97.7\% and the low-level improves from 84.3\% to 86\%.
Further increasing the number of self-correction turns can further improve the model performance.
For a total of 4 reflection turns, it achieves 86.5\% low-level score and 84.9\% high-level score.

\noindent\textbf{Deep dive into self-correction ability.}
To further understand the self-correction behavior of each model, we compute the following metrics on ChartMimic in \Cref{tab:self_correction_detailed_metrics}.
\begin{itemize}
    \item \textbf{Average score improvement} on charts whose first- and second-turn solutions are both executable.
    \item \textbf{Percentage of improved samples} between two turns. Due to the stochasticity of the high-level score, we calculate the percentage of charts whose high-level scores are improved by at least 10\%.
    \item \textbf{Percentage of degraded samples}, similar to the above but for score degradation.
\end{itemize}
We find that while Qwen2.5-VL and Qwen3-VL raise the scores for a subset of charts, a comparable proportion of samples exhibit score degradation, leading to an overall gain that is near zero or even negative.
On the contrary, MM-ReCoder improves $7\%$ of the charts while only $4\%$ are degraded, resulting in $0.3\%$ improvement over low-level score and $0.89\%$ improvement over high-level score compared with charts rendered in the first turn.

Beyond two turns, we evaluate the model for up to five turns to assess whether it can continue improving image quality. As shown in \Cref{tab:iterative_self_correct}, MM-ReCoder improves its outputs over the first few turns, but both the proportion of improved samples and the magnitude of score gains gradually diminish. By the fifth turn, the model is no longer able to further increase the low-level score.

\begin{figure}[t]
    \centering
    \vspace{-10pt}
    \begin{subfigure}[b]{0.95\linewidth}
        \centering
        \includegraphics[width=\linewidth]{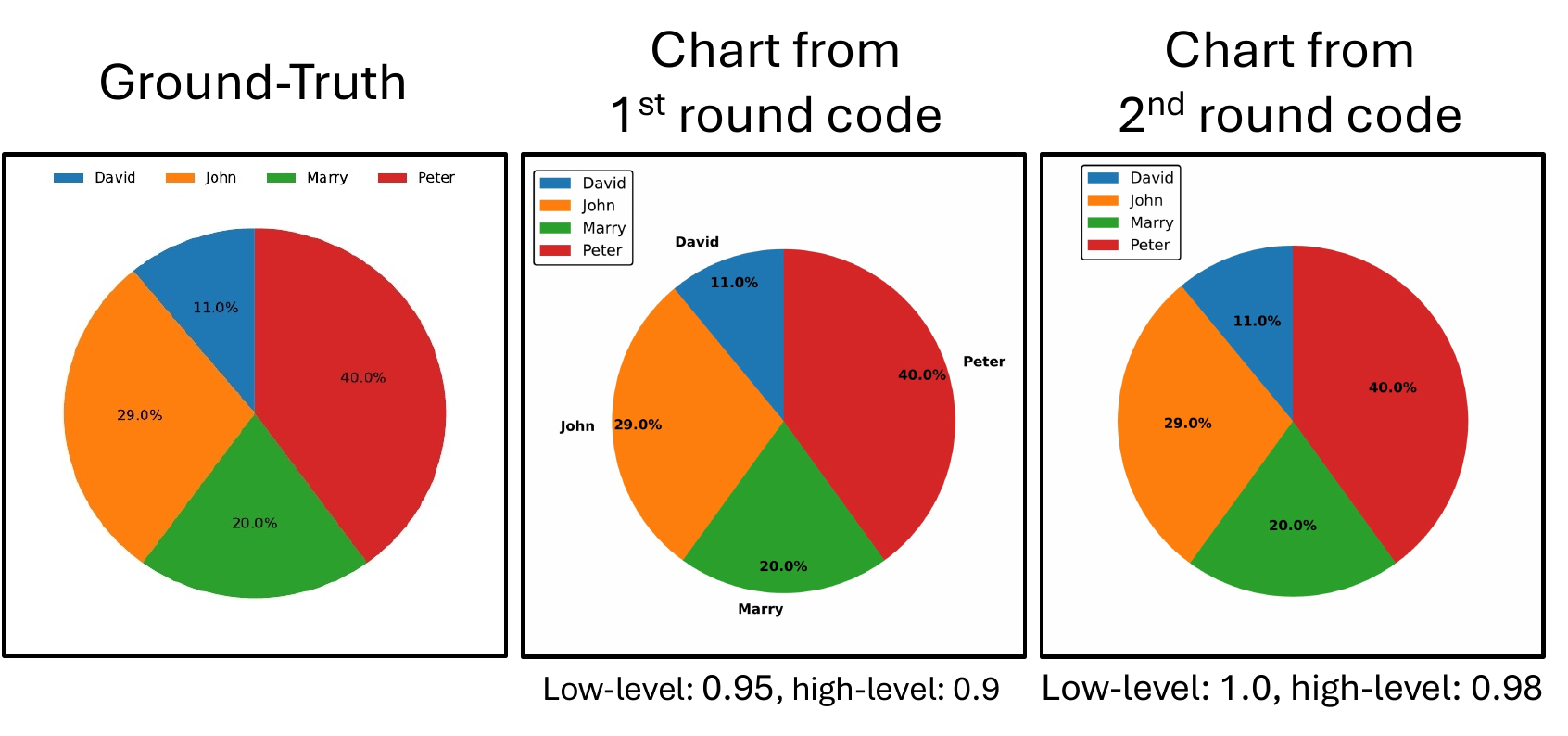}
        \vspace{-15pt}
        \caption{The $1^{st}$-round chart captures the overall structure and proportions but introduces label placement shifts, while the $2^{nd}$-round correction refines these stylistic details to more closely match the ground-truth.}
    \end{subfigure}
    
    \begin{subfigure}[b]{0.95\linewidth}
        \centering
        \includegraphics[width=\linewidth]{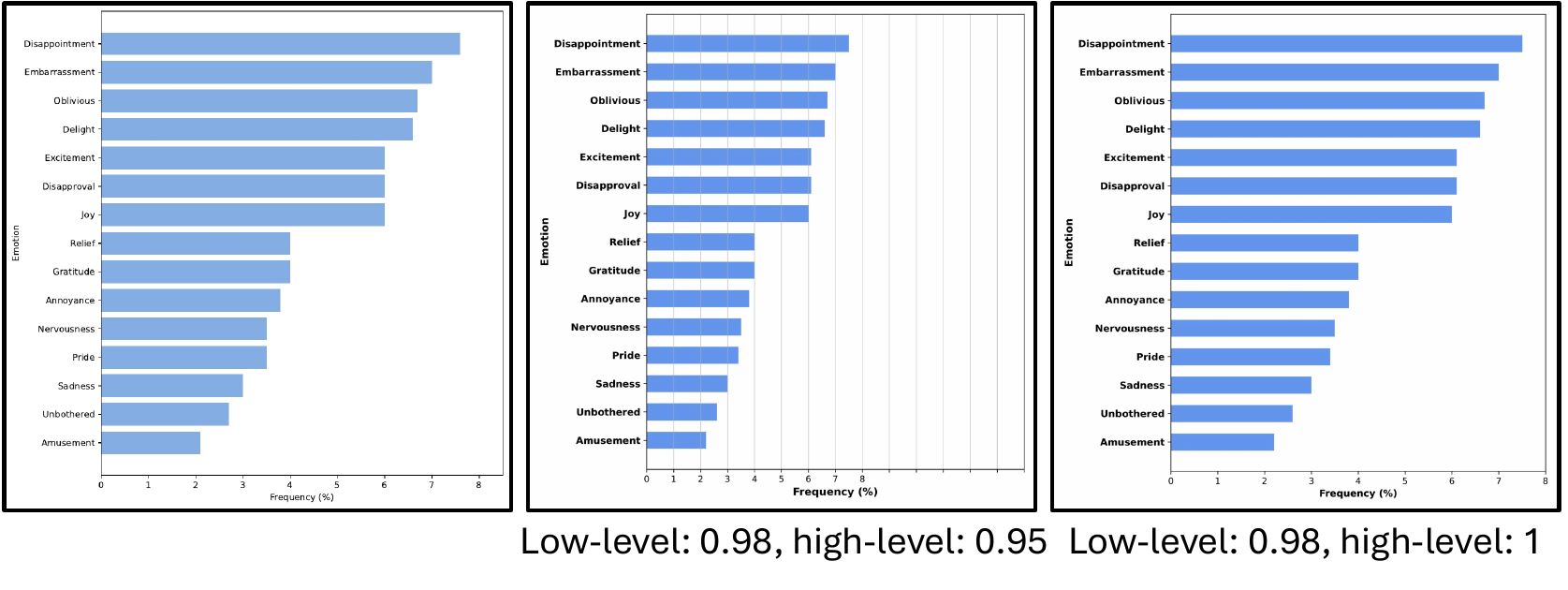}
        \vspace{-15pt}
        \caption{The $1^{st}$-round output preserves the overall ordering and bar magnitudes but exhibits misaligned axes and spacing. The $2^{nd}$-round self-correction refines these details, matching ground-truth more closely.}
    \end{subfigure}
    
    \begin{subfigure}[b]{0.95\linewidth}
        \centering
        \includegraphics[width=\linewidth]{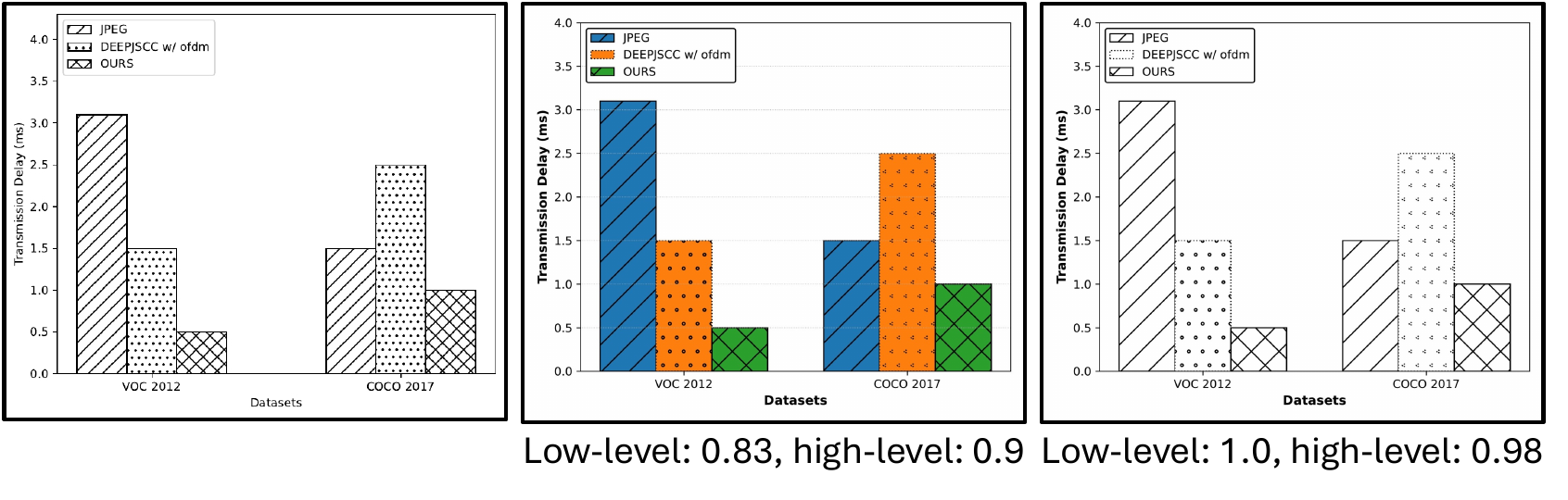}
        \vspace{-15pt}
        \caption{The $1^{st}$-round output matches the grouping and bar heights but introduces color-style mismatches. The $2^{nd}$-round self-correction restores the correct colors and improves alignment with the ground-truth.}
    \end{subfigure}
    \vspace{-5pt}
    \caption{Qualitative results comparing ground truth, $1^{st}$-round generation, and $2^{nd}$-round self-correction (left to right).}
    \vspace{-15pt}
    \label{fig:qualitative}
\end{figure}

\noindent\textbf{Qualitative results.} We illustrate self-correction examples of MM-ReCoder in~\Cref{fig:qualitative}. Our model can revise the rendered texts, axis ranges, chart styles, \emph{etc.} between two turns.

\subsection{Analysis of Self-correction Training Strategies}
\label{sec:abla_self_correction}

We compare all three RL training strategies introduced in~\Cref{sec:training_strategy}, namely, full-trajectory optimization, shared first turn, and two-stage training.
For the full-trajectory optimization, we also explore different first turn reward factor $\gamma$ and self-correction bonus $\eta$. In addition, we evaluate a turn-wise training strategy in which, instead of applying shared-first-turn across all samples, 50\% of the samples are used to perform single-turn RL without self-correction. This enables the model to learn both first- and second-turn behavior.

To accelerate experimentation, we use only the format reward and rule-based reward, excluding the model-based reward.
We evaluate whether their self-correction can improve the low-level score on ChartMimic~\cite{yang2024chartmimic}.

The results are in~\Cref{tab:compare_strategies}.
All the strategies can improve the execution rate.
However, when looking into the low-level score, the model behaviors are different.
Under full-trajectory optimization with $\gamma=\eta=0$, the model repeats its first-turn solution in the second turn in 46.9\% of the cases and only 3.4\% of charts show improvement.
Introducing a self-correction bonus $\eta=0.1$ causes the model to exploit the reward by intentionally producing extremely poor first-turn outputs.
Setting $\gamma=0.5$ partially stabilizes this behavior by allowing the first-turn score to contribute to the final reward. However, its self-correction capability, indicated by its average low-level improvement of $-0.03\%$ and code repetition rate of $55.6\%$, is diminished.

We attribute the code repetition behavior of the full-trajectory optimization to the incapability of the base model in self-correction.
As shown in~\Cref{tab:self_correction_of_cold_start}, the average score improvement is negative even after cold start.
As a result, when the first turn output is executable, the expected reward gain of generating a different solution is negative, which encourages the model to repeat its solution.
This highlights the need to disentangle the training of the two turns.

\begin{table}[t]
\setlength{\tabcolsep}{3pt}
\small
\centering
\resizebox{.48\textwidth}{!}{
\begin{tabular}{c|cccc}
\toprule
\multirow{2}{*}{Strategy} & \makecell{First turn \\ low-level} & \makecell{Second turn \\ low-level} & \makecell{Avg. improvement \\ on rendered charts} & \makecell{Percentage of \\ repeated code} \\
\midrule
Qwen2.5-VL-7B & 56.2 & 57.2 & -0.36 & 10.9 \\
+ Single-turn cold start & 79.1 & 78.9 & -0.10 & 81.5 \\
+ Multi-turn cold start & 75.2 & 73.5 & -0.54 & 2.3 \\
+ RL & 83.5 & 84.8 & 0.30 & 2.2 \\
\bottomrule 
\end{tabular}}
\vspace{-5pt}
\caption{Performance after each stage. Cold start improves coding ability, but cannot directly enable self-correction. Multi-turn cold start recovers the model's multi-turn ability but degrades the performance because the data in this stage is not ground truth.}
\label{tab:self_correction_of_cold_start}
\vspace{-5pt}
\end{table}

\begin{table}[t]
\centering
    \resizebox{0.48\textwidth}{!}{
    \begin{tabular}{ccc|ccc}
        \toprule
        \multicolumn{3}{c|}{Reward weights} & \multicolumn{3}{c}{ChartMimic} \\
        \toprule
        Format & Rule-based & Model-based & Exec.Rate & Low-level & High-level \\
        \cmidrule(r){1-6}
        0.1 & 0.9 & 0.0 & 95.0 & 84.8 & 78.6 \\
        0.1 & 0.8 & 0.1 & 95.0 & 84.3 & 83.7 \\
        0.1 & 0.6 & 0.3 & 94.3 & 84.1 & 83.7 \\
        0.1 & 0.4 & 0.5 & 95.5 & 84.2 & 85.3 \\
        0.1 & 0.2 & 0.7 & 93.3 & 83.9 & 83.8 \\
        0.1 & 0.0 & 0.9 & 95.5 & 78.2 & 83.8 \\
        \bottomrule
    \end{tabular}
    }
\vspace{-5pt}
\caption{Ablation on RL reward weights. Model training and inference are single-turn without self-correction.}
\label{tab:reward_ablation}
\vspace{-15pt}
\end{table}

With this in mind, we evaluate turn-wise training and shared-first-turn.
For turn-wise training, the model repeats its first-turn output in 63\% of samples and shows minimal self-correction capability (average improvement of just 0.05\%). In contrast, shared-first-turn demonstrates strong self-correction ability, despite slightly lower total low-level scores. It improves the chart in 14.4\% of cases, with an average low-level improvement of 
0.72\%.

The proposed two-stage training first uses second-turn-only optimization to elicit the model's self-correction ability and then employs full-trajectory optimization to enhance its overall coding skill.
The resulting model retains strong self-correction ability while achieving the highest low-level score (86\%) among all strategies. Further introducing $\gamma$ and $\eta$ in the second stage does not improve the performance.

\subsection{Analysis of RL Rewards}
\label{sec:reward_weight}
We conduct ablation study on the effectiveness of the rule-based reward and the model-based reward in RL training.
For clarity, we fix the format reward weight to 0.1 throughout our experiments.
Results in~\Cref{tab:reward_ablation} show that, by removing the model-based reward, the high-level score of the model drops significantly from 83.7\% to 78.6\%, causing the model to generate charts with poor clarity (\Cref{fig:reward_hacking}).
By removing the rule-based score, the low-level score of the model drops from 83.9\% to 78.2\%, indicating unfaithfully reproduced visual elements.
Interestingly, we observe that the exact weight of each reward component has limited impact on the final performance.
Even a small reward weight provides sufficient signal for the model to fit, and enlarging the weight cannot facilitate the learning of the reward.

%% file: sec/5_conclusions.tex
\section{Conclusions}
\label{sec:conclusions}

In this paper, we introduce MM-ReCoder, the first chart-to-code generation model equipped with self-correction capability.
To enable self-correction, we propose a two-stage self-correction RL strategy.
In the first stage, we roll out a shared first turn and explicitly require the model to self-correct in the second turn.
In the second stage, we perform full-trajectory optimization to jointly improve the model's coding capability in both turns.
Our results on three benchmarks show that MM-ReCoder achieves the best performance among chart-domain models and models of comparable size.
It even outperforms state-of-the-art MLLMs on some metrics.
MM-ReCoder shows better self-correction ability than all open-source MLLMs that we evaluated.

%% file: sec/X_suppl.tex
\clearpage
\setcounter{page}{1}
\setcounter{table}{0}
\renewcommand{\thetable}{A\arabic{table}}
\setcounter{figure}{0}
\renewcommand{\thefigure}{A\arabic{figure}}
\maketitlesupplementary

\appendix

We elaborate on the details of our rewards and MM-ReCoder prompts in~\Cref{app:implementation}.
\Cref{app:curves} shows our RL training curves.
In \Cref{app:ablation}, we conduct ablation studies on the RL training strategy, cold start, and the model used in the model-based reward.
We further compare MM-ReCoder with a few baselines through human evaluation in \Cref{app:human_eval}.
Finally, qualitative results are in \Cref{app:qualitative}.

\section{Implementation Details}
\label{app:implementation}

In this section, we first provide a detailed description of the prompt used for the model-based reward.
Then we elaborate on the prompt we use to guide MM-ReCoder in self-correction.

\subsection{Reward Implementation}

In the rule-based reward, we evaluate text score, color score, chart type score, and layout score of the generated chart.
The implementation of all these scores is borrowed from ChartMimic~\cite{yang2024chartmimic}.
Please refer to their paper and code for implementation.

To measure the model-based reward, we use vLLM to launch a Qwen2.5-VL-72B server and set the sampling temperature to be $0.3$.
If the generated code encounters runtime error, its model-based reward is 0.
Otherwise, we use the following prompt to ask Qwen2.5-VL-72B to score the generated chart:
\begin{tcolorbox}[breakable,colback=green!5!white, colframe=green!50!black]
\textcolor{red}{\textless ground truth image\textgreater\quad} \\
\textcolor{red}{\textless generated image\textgreater\quad} \\
You are an excellent judge at evaluating visualization chart plots.
The first image (reference image) is created using ground truth matplotlib code, and the second image (AI-generated image) is created using matplotlib code generated by an AI assistant. Your task is to score how well the AI-generated plot matches the ground truth plot. \\
\\
\textcolor{blue}{\#\#\# Scoring Methodology:} \\
The AI-generated image's score is based on the following criteria, totaling a score out of 100 points: \\
1. \textbf{Chart Types (20 points)} Does the AI-generated image include all chart types present in the reference image (e.g., line charts, bar charts, etc.)?\\
2. \textbf{Layout (10 points)} Does the arrangement of subplots in the AI-generated image match the reference image (e.g., number of rows and columns)?\\
3. \textbf{Text Content (20 points)} Does the AI-generated image include all text from the reference image (e.g., titles, annotations, axis labels), excluding axis tick labels?\\
4. \textbf{Data (20 points)} How accurately do the data trends in the AI-generated image resemble those in the original image and is the number of data groups the same as in the reference image?\\
5. \textbf{Style (20 points)} Does the AI-generated image match the original in terms of colors (line colors, fill colors, etc.), marker types (point shapes, line styles, etc.), legends, grids, and other stylistic details?\\
6. \textbf{Clarity (10 points)} Is the AI-generated image clear and free of overlapping elements?\\
\\
\textcolor{blue}{\#\#\# Evaluation:}\\
Compare the two images head to head and provide a detailed assessment. Use the following format for your response:\\
---\\
Detailed Assessment: \$\{your assessment for each point\}\\
---\\
Scores:\\
- \textbf{Chart Types:} \$\{your score out of 20\}\\
- \textbf{Layout:} \$\{your score out of 10\}\\
- \textbf{Text Content:} \$\{your score out of 20\}\\
- \textbf{Data:} \$\{your score out of 20\}\\
- \textbf{Style:} \$\{your score out of 20\}\\
- \textbf{Clarity:} \$\{your score out of 10\}\\
\\
\textbf{Total Score:} \$\{your total score out of 100\}\\
---\\
\\
Please use the above format to ensure the evaluation is clear and comprehensive.
\end{tcolorbox}

\subsection{Model Prompt of MM-ReCoder}

In the first turn of the conversation, we give MM-ReCoder a chart image and ask it to reproduce it using Python.
We also include an instruction to follow the template requirement of our format reward described in \Cref{sec:reward_design}.
Specifically, our prompt is:

\begin{tcolorbox}[breakable,colback=green!5!white, colframe=green!50!black]
\textcolor{red}{\textless image\textgreater\quad} \\
You are an expert developer specializing in writing Python matplotlib code based on a given picture. I need your help to generate the Python code that can reproduce the picture based on the picture I provided.\\
\textcolor{blue}{You FIRST think about the reasoning process as an internal monologue and then provide the final code. The reasoning process MUST BE enclosed within \textless think\textgreater \textless/think\textgreater\; tags. The final code MUST BE put in \\
\texttt{```}python\\
Your code\\
\texttt{```}\\
at the end.}
\end{tcolorbox}

Notice that the specific task prompt, which is the black text in the prompt above, may differ from dataset to dataset.
The one we show here is an example from Chart2Code-160k~\cite{zhao2025chartcoder}.
During training and evaluation, we use the task prompt provided by each dataset.

After the model outputs a first-turn response, we extract and execute the code inside the \texttt{```python...```} block.
If the execution encounters a runtime error, we get the error information from the Python interpreter.
Otherwise, we derive an image rendered from the code.

In the second turn, if the feedback is a runtime error, we instruct the model to revise its code as the following:
\begin{tcolorbox}[breakable,colback=green!5!white, colframe=green!50!black]
\textcolor{red}{\textless error information\textgreater\quad} \\
Your code encountered a runtime error. Please continue modifying the code.\\
\textcolor{blue}{You FIRST think about the reasoning process as an internal monologue and then provide the final code. The reasoning process MUST BE enclosed within \textless think\textgreater \textless/think\textgreater\; tags. The final code MUST BE put in \\
\texttt{```}python\\
Your code\\
\texttt{```}\\
at the end.}
\end{tcolorbox}

If the chart is successfully rendered, we use the following prompt:
\begin{tcolorbox}[breakable,colback=green!5!white, colframe=green!50!black]
\textcolor{red}{\textless image\textgreater\quad} \\
Here is the image rendered by your code. Please continue modifying the code so that the reproduced chart looks more like the given image.\\
\textcolor{blue}{You FIRST think about the reasoning process as an internal monologue and then provide the final code. }\textcolor{blue}{The reasoning process MUST BE enclosed within \textless think\textgreater \textless/think\textgreater\; tags. The final code MUST BE put in} \\
\textcolor{blue}{\texttt{```}python\\
Your code\\
\texttt{```}\\
at the end.}
\end{tcolorbox}

When we employ MM-ReCoder for more than two turns, we repeat this process and the conversations in the previous turns are all kept as contexts.

\section{RL Training Curves}
\label{app:curves}

\begin{figure*}[t]
    \centering

    \begin{subfigure}{0.4\textwidth}
        \centering
        \includegraphics[width=\linewidth]{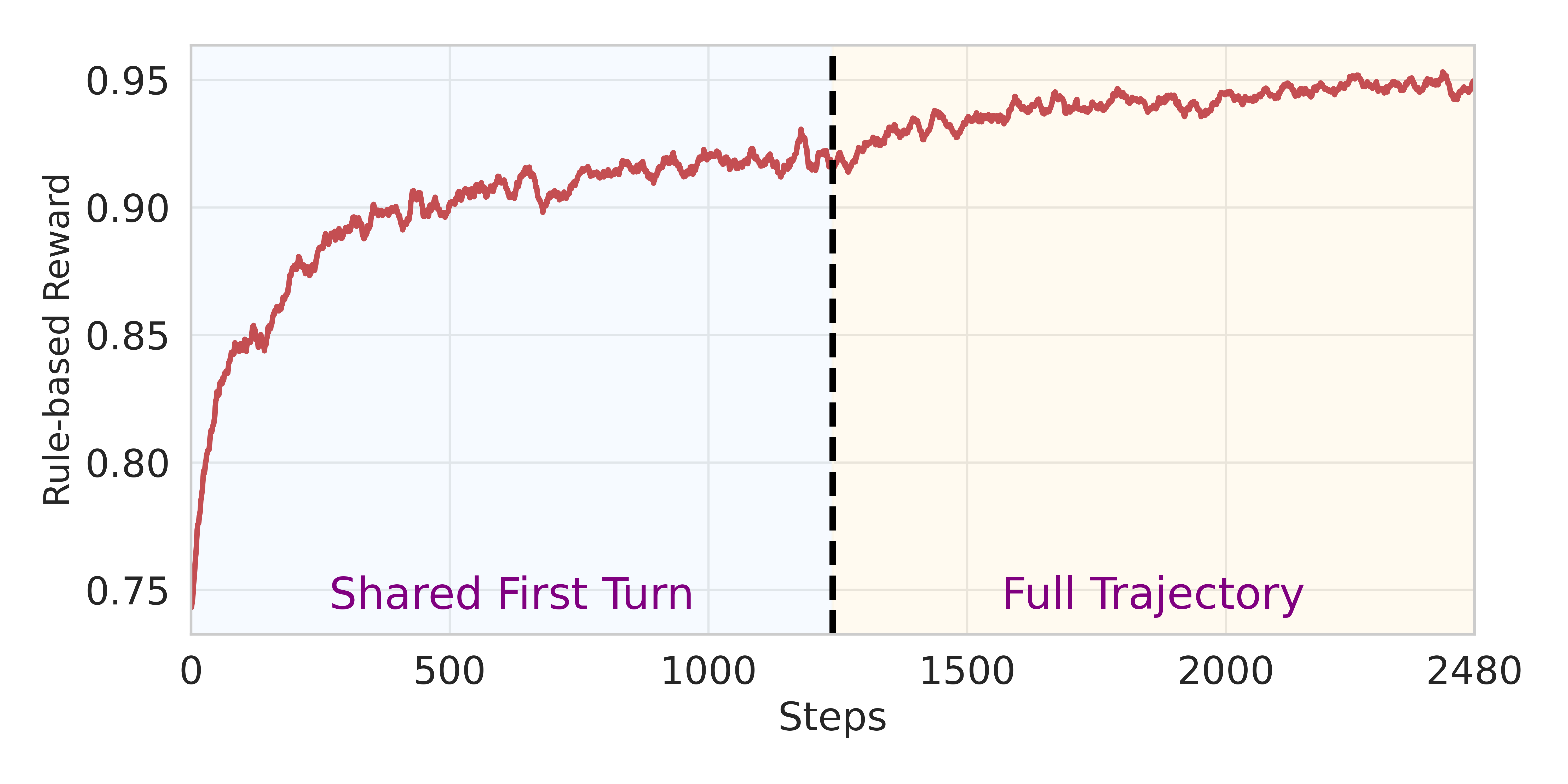}
        \caption{Rule-based Reward during Training}
    \end{subfigure}
    \begin{subfigure}{0.4\textwidth}
        \centering
        \includegraphics[width=\linewidth]{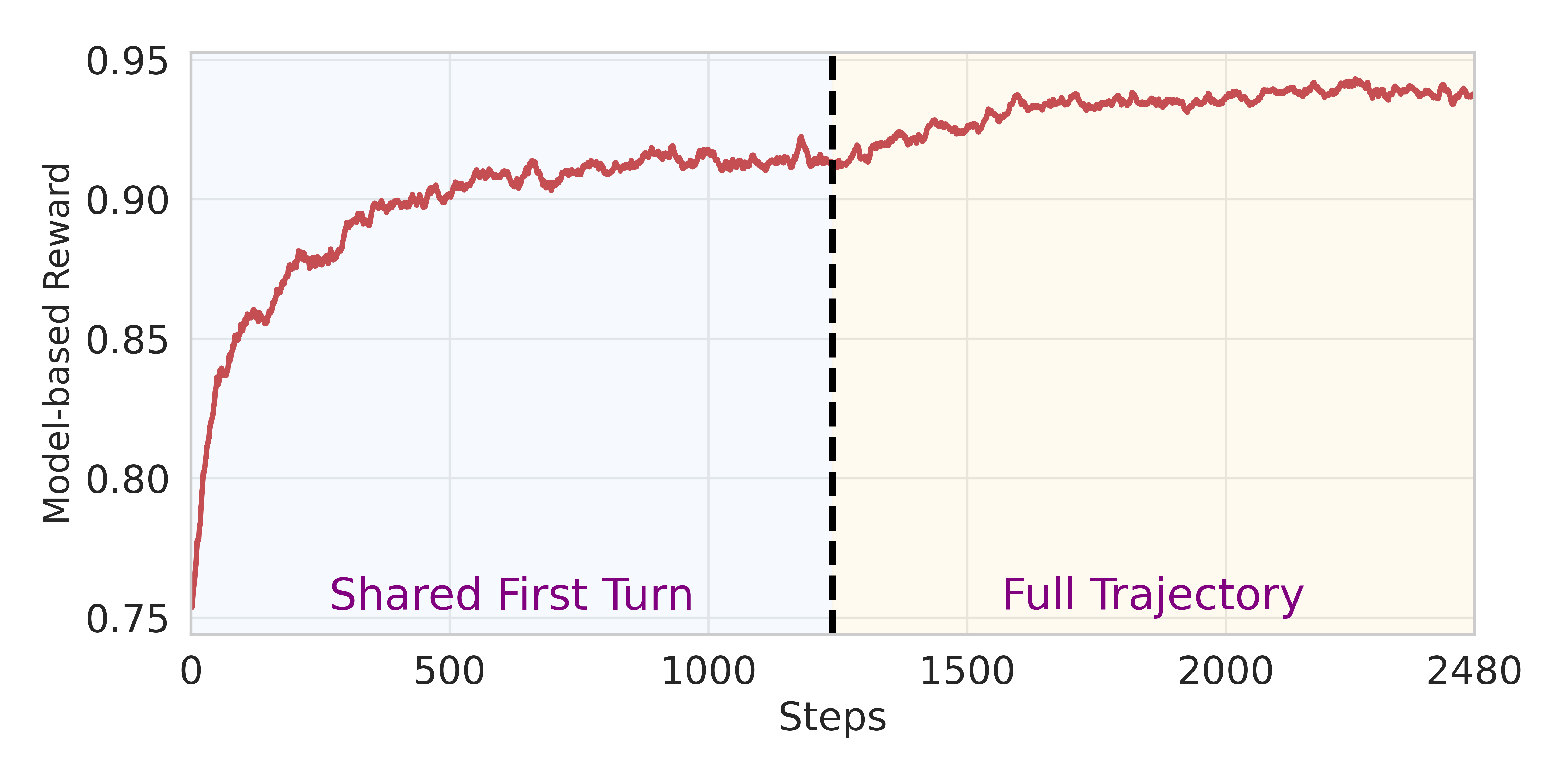}
        \caption{Model-based Reward during Training}
    \end{subfigure}
    
    \vspace{0.3cm}
    
    \begin{subfigure}{0.4\textwidth}
        \centering
        \includegraphics[width=\linewidth]{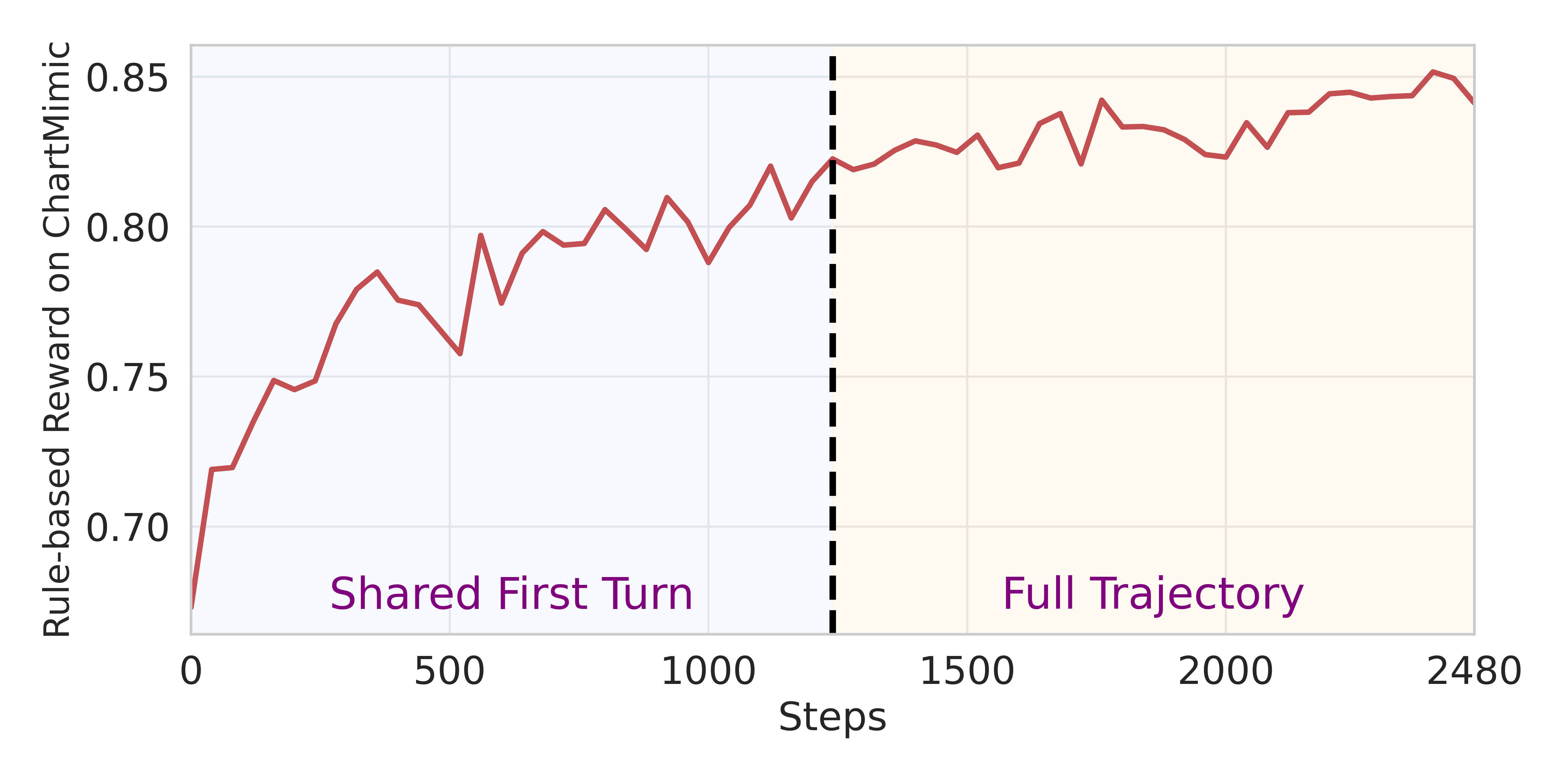}
        \caption{Rule-based Reward on ChartMimic}
    \end{subfigure}
    \begin{subfigure}{0.4\textwidth}
        \centering
        \includegraphics[width=\linewidth]{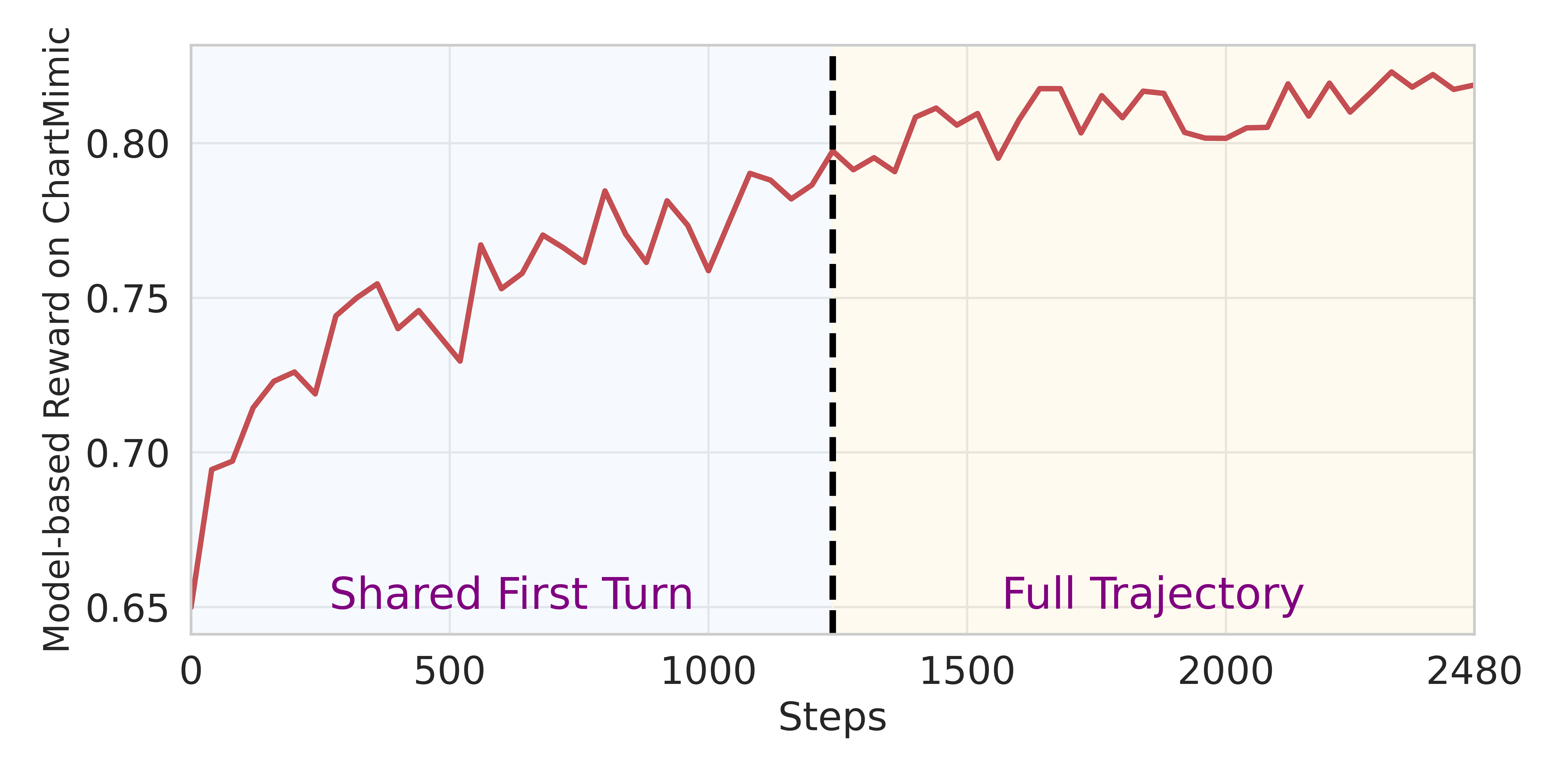}
        \caption{Model-based Reward on ChartMimic}
    \end{subfigure}
    
    \vspace{0.3cm}
    
    \begin{subfigure}{0.4\textwidth}
        \centering
        \includegraphics[width=\linewidth]{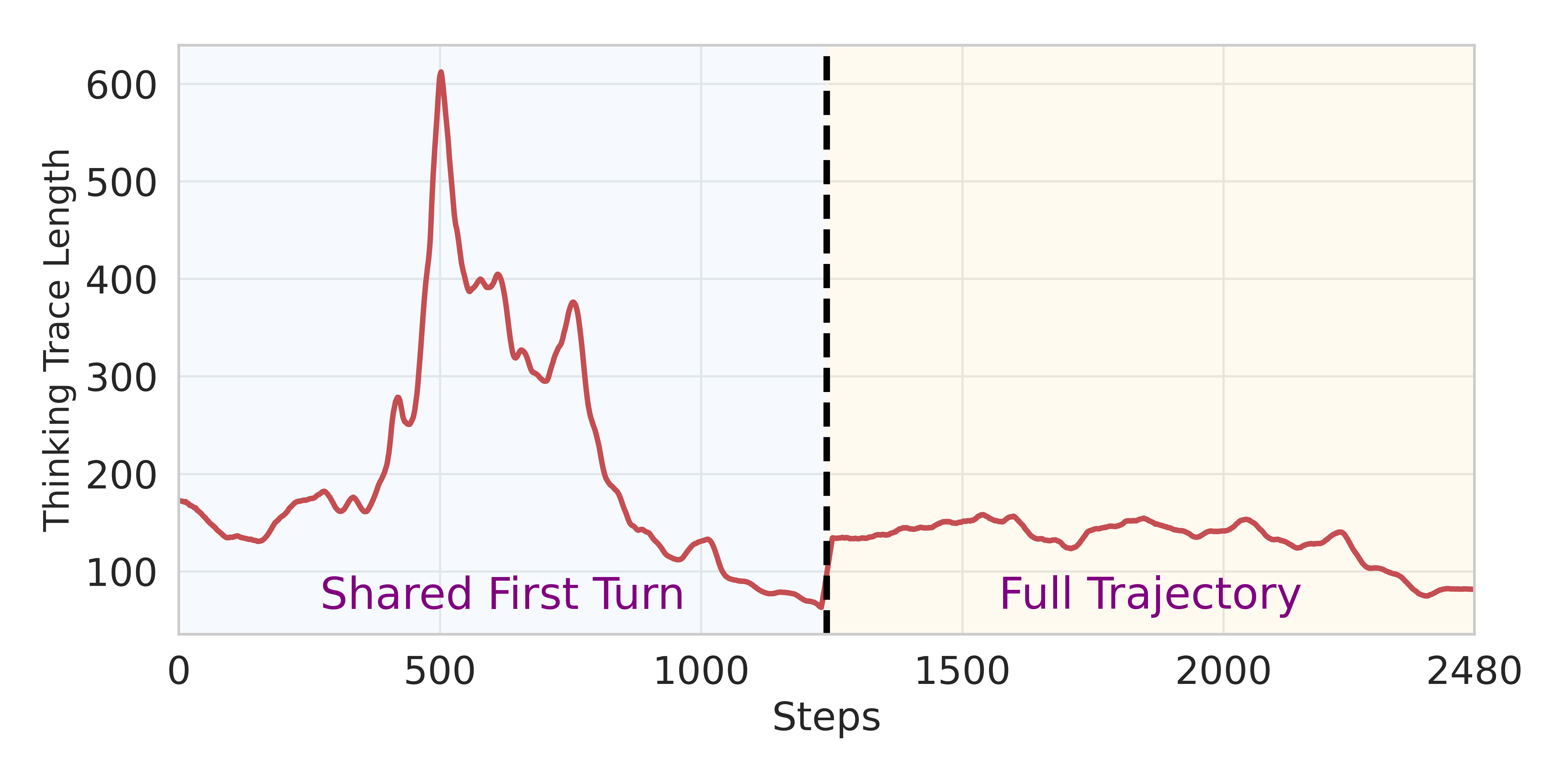}
        \caption{Length of Thinking Trace during Training}
    \end{subfigure}
    \begin{subfigure}{0.4\textwidth}
        \centering
        \includegraphics[width=\linewidth]{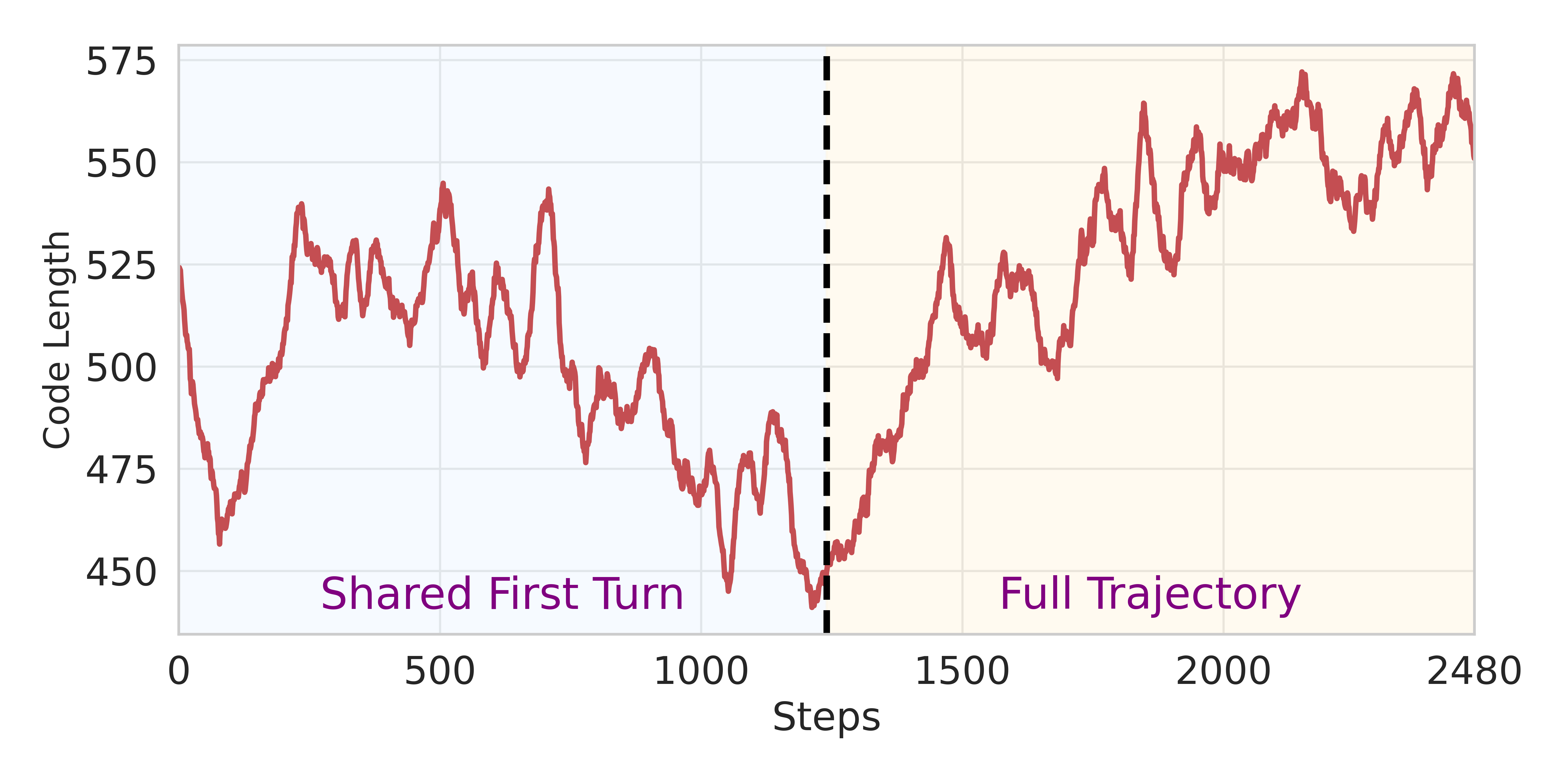}
        \caption{Length of Code during Training}
    \end{subfigure}

    \caption{Training curves of MM-ReCoder. Curves on the training set are smoothed with a window size of 20 steps. We evaluate the model on ChartMimic every 40 steps. Note that the value of the rule-based reward can be slightly different from the reported model performance on low-level score because the latter is evaluated under the official evaluation codebase of ChartMimic.}
    \label{fig:training_curves}
\end{figure*}

In \Cref{fig:training_curves}, we show the training reward curves and the length of thinking trace/code of MM-ReCoder.
We also evaluate the model every 40 steps on ChartMimic during training.

We find that the training rewards reach a plateau at the end of the first RL stage --- shared first turn.
But they begin to rise again in the second stage, full-trajectory optimization.
On the contrary, the evaluation rewards on ChartMimic keep rising during training.
However, the evaluation rewards can only hit $85\%$ while the training rewards reach $95\%$.
This shows that there is a distribution shift between the training data and the evaluation benchmarks.

One interesting observation is that, during the shared-first-turn stage, the length of the model's thinking trace first grows from $\sim200$ tokens to $\sim600$ tokens, and then gradually drops below 100 tokens.
During the first half of the shared-first-turn stage, we find the model learns to propose a long item list of code revision.
At the 500th step, the model includes roughly 30 revision items in the thinking trace, leading to an average length of 600 tokens.
After this point, the model gradually discards incorrect revision items and finally converges to around four items per chart.

\section{Ablation Studies}
\label{app:ablation}

In this section, we conduct ablation studies on the self-correction RL training strategy, cold start, and the model used in the model-based reward.

\begin{table*}[t]
\small
\centering
\resizebox{\textwidth}{!}{
\begin{tabular}{c|cc|cc|cc|cccc}
\toprule
\multirow{2}{*}{Strategy} & \multirow{2}{*}{$\gamma$} & \multirow{2}{*}{$\eta$} & \multicolumn{2}{c|}{First turn} & \multicolumn{2}{c|}{Second turn} &  \multirow{2}{*}{\makecell{Avg. low-level improvement \\ on rendered charts}} & \multirow{2}{*}{\makecell{Percentage of \\ improved samples}} & \multirow{2}{*}{\makecell{Percentage of \\ degraded samples}} & \multirow{2}{*}{\makecell{Percentage of \\ repeated code}} \\
\cmidrule{4-7} 
  & & & Exec.Rate & Low-Level & Exec.Rate & Low-Level &  \\
\midrule
Full-trajectory & 0   & 0   & 91.8 & 81.8 & 95.2 & 83.9 & 0.21 & 3.4 & 2.6 & 46.9 \\
\midrule
                & 0   & 0.1 & 64.5 & 66.7 & 96.0 & 84.3 & 10.11 & 87.5 & 2.3 & 0.3 \\
Full-trajectory & 0   & 0.3 & 79.0 & 73.7 & 94.5 & 83.2 & 4.57 & 77.1 & 8.3 & 0.0 \\
                & 0   & 0.5 & 71.8 & 72.5 & 96.2 & 77.7 & 3.13 & 77.7 & 10.2 & 0.0 \\
\midrule
                & 0.5 & 0.1 & 96.5 & 85.0 & 97.5 & 85.5 & -0.03 & 2.2 & 2.2 & 55.6 \\
Full-trajectory & 0.8 & 0.1 & 95.2 & 83.7 & 96.3 & 84.9 & 0.01 & 3.2 & 3.9 & 45.9 \\
                & 1 & 0.1 & 95.5 & 84.7 & 96.2 & 84.2 & 0.00 & 1.4 & 1.4 & 5.1 \\
\bottomrule 
\end{tabular}}
\vspace{-5pt}
\caption{Ablation on the hyperparameters in full-trajectory optimization. All experiments are conducted \textbf{without model-based reward}.}
\label{tab:ablation_hyper_full}
\vspace{-15pt}
\end{table*}

\begin{table*}[t]
\small
\centering
\resizebox{\textwidth}{!}{
\begin{tabular}{c|cc|cc|cc|cccc}
\toprule
\multirow{2}{*}{Strategy} & \multirow{2}{*}{$\gamma$} & \multirow{2}{*}{$\eta$} & \multicolumn{2}{c|}{First turn} & \multicolumn{2}{c|}{Second turn} &  \multirow{2}{*}{\makecell{Avg. low-level improvement \\ on rendered charts}} & \multirow{2}{*}{\makecell{Percentage of \\ improved samples}} & \multirow{2}{*}{\makecell{Percentage of \\ degraded samples}} & \multirow{2}{*}{\makecell{Percentage of \\ repeated code}} \\
\cmidrule{4-7} 
  & & & Exec.Rate & Low-Level & Exec.Rate & Low-Level &  \\
\midrule
\multirow{2}{*}{Full-trajectory} & 0   & 0   & 91.8 & 81.8 & 95.2 & 83.9 & 0.21 & 3.4 & 2.6 & 46.9 \\
                & 0.5 & 0.1 & 96.5 & 85.0 & 97.5 & 85.5 & -0.03 & 2.2 & 2.2 & 55.6 \\
\midrule
\multirow{2}{*}{\makecell{Full-trajectory \\ for 2 epochs}} & 0   & 0   & 92.3 & 79.0 & 97.0 & 81.8 & 0.16 & 4.5 & 2.5 & 54.0 \\
                & 0.5 & 0.1 & 94.3 & 82.2 & 94.2 & 82.3 & 0.07 & 2.7 & 1.6 & 87.7 \\
\midrule
Shared-first-turn & -  & -   & 91.7 & 79.8 & 95.8 & 82.6 & 0.72 & 14.4 & 8.4 & 16.8 \\
\midrule
\makecell{Shared-first-turn \\ for 2 epochs} & -   & -  & 92.3 & 82.3 & 97.2 & 84.8 & 0.35 & 10.5 & 4.7 & 12.3 \\
\midrule
\makecell{Shared-first-turn \\ + Full-trajectory} & 0   & 0   & 94.3 & 83.7 & 97.7 & \textbf{86.0} & \textbf{0.55} & 12.1 & 8.2 & 21.6 \\
\midrule
\makecell{Full-trajectory \\ + Shared-first-turn} & 0   & 0   & 93.3 & 83.6 & 97.2 & 85.0 & 0.02 &  2.9 & 2.7 & 46.7 \\
\midrule
\makecell{Single-turn \\ + Full-trajectory} & 0   & 0   &  94.7 & 84.5 & 96.5 & 85.2 & 0.03 & 2.7 & 4.0 & 58.4 \\
\bottomrule 
\end{tabular}}
\vspace{-5pt}
\caption{Ablation on the RL stage combinations. All experiments are conducted \textbf{without model-based reward}.}
\label{tab:ablation_stage_comb}
\vspace{-3pt}
\end{table*}

\begin{table*}[t]
\setlength{\tabcolsep}{3pt}
\small
\centering
\resizebox{.75\textwidth}{!}{
\begin{tabular}{c|cccc}
\toprule
\multirow{3}{*}{Cold Start} & \multicolumn{4}{c}{Performance on ChartMimic after RL} \\
\cmidrule{2-5}
& \makecell{First turn \\ low-level} & \makecell{Second turn \\ low-level} & \makecell{Average improvement \\ on rendered charts} & \makecell{Percentage of \\ repeated code} \\
\midrule
None & 83.0 & 84.9 & 0.53 & 36.3 \\
Single-turn cold start & 82.3 & 84.8 & 0.02 & 13.1 \\
\makecell{Single-turn cold start + multi-turn cold start} & 83.7 & 86.0 & 0.55 & 21.6 \\
\bottomrule 
\end{tabular}}
\vspace{-5pt}
\caption{Ablation studies on cold start. We train the base model with different cold start recipes and the same RL strategy (full-trajectory + shared-first-turn). We evaluate their performance on ChartMimic after RL. Experiments are conducted \textbf{without model-based reward}.}
\label{tab:ablation_cold_start}
\vspace{-15pt}
\end{table*}

\subsection{RL Training Strategy}

\noindent\textbf{Hyperparameters in full-trajectory optimization.}
In our full-trajectory optimization method, we use a hyperparameter $\gamma$ to balance the rewards of the two turns and a hyperparameter $\eta$ to encourage the model for successful self-correction.
In \Cref{sec:abla_self_correction}, we claim that $\gamma=0.5$ and $\eta=0.1$ work best when only using the full-trajectory optimization.
We perform ablation studies on this design choice in \Cref{tab:ablation_hyper_full}.

Firstly, when the reward of the first turn is not utilized ($\gamma=0$), the model always lowers the quality of its first-round output to hack the self-correction bonus $\eta$.
And higher $\eta$ leads to worse second-turn performance.
Secondly, when $\eta=0.1$, the existence of $\gamma$ can prevent the model from lowering the first-turn quality and achieve better performance.
However, it also hinders the model from learning self-correction, indicated by the nearly-zero low-level score improvement on rendered charts.
A $\gamma$ of $0.5$ shows the best overall performance across all the choices we have tried.

\noindent\textbf{RL strategy combinations.}
While our RL strategy in MM-ReCoder is shared-first-turn optimization followed by full-trajectory optimization, we try out several other possible combinations of the RL strategies in \Cref{tab:ablation_stage_comb}.

As our method is composed of two stages, which trains the model for two epochs on the training set in total, we first try employing the full-trajectory and shared-first-turn optimizations for two epochs for fair comparison. For the full-trajectory optimization, however, training with it for two epochs causes over-fitting, achieving lower performance than one epoch.
On the other hand, shared-first-turn for two epochs can further boost the model performance compared to one epoch, but underperforms the other strategies.

We explore strategy combinations besides the one we use in our main method.
For example, we can switch the order of the two stages, performing full-trajectory optimization followed by shared-first-turn.
Moreover, we can also train a model under the single-turn RL setting first and then perform full-trajectory optimization to enable self-correction.
However, as shows in \Cref{tab:ablation_stage_comb}, these two strategies cannot elicit the self-correction capability of the model and result in suboptimal performance.
They repeat their first-turn code in the second turn with a frequency of nearly $50\%$.
We attribute this to their weak self-correction ability after full-trajectory optimization and single-turn RL.
In this case, the model’s output entropy is significantly reduced during the first-stage RL, making it unable to explore self-correction effectively in the second stage.

\subsection{Cold Start}

To investigate the effect of cold start, we perform our two-stage RL on two checkpoints: (a) Qwen2.5-VL-7B without any cold start and (b) the checkpoint that has only undergone the single-turn cold start.
The results are in \Cref{tab:ablation_cold_start} and the RL stage is performed without model-based reward.

We find that our self-correction RL training strategy can enable Qwen2.5-VL-7B to self-correct without any cold start, indicating that the teacher model used to generate the code start data is not the key to elicit self-correction.
However, the performance without cold start is around $1\%$ lower than that of the cold-started model.
While cold start on the single-turn data, Chart2Code-160k, can improve the base model's coding capability as shown in \Cref{tab:self_correction_of_cold_start}, directly applying RL on this checkpoint does not realize self-correction because of its weak capability in multi-turn conversation.
After the multi-turn cold start stage, the model is able to benefit from both the cold start and our multi-turn self-correction RL method.

We also look into the qualitative results to investigate the effect of cold start. 
We notice that the model without cold start generates very short and abstract thinking traces, while the one with cold start learns to describe the image differences in detail and outlines an action list before coding,
which is important to boost the model's performance.

\subsection{Model-based Reward}

In the RL reward of MM-ReCoder, we use Qwen2.5-VL-72B judger as a model-based reward to score the generated charts.
While Qwen2.5-VL-72B is larger than our base model, Qwen2.5-VL-7B, we show in \Cref{tab:reward_model_ablation} that using Qwen2.5-VL-7B itself as the reward model can also guide the learning of MM-ReCoder.

In this experiment, we train MM-ReCoder in the single-turn setting without self-correction.
We find that using Qwen2.5-VL-7B as the reward model can improve the high-level score by $4\%$ compared to training without model-based reward.
However, the model's low-level score is degraded by $3.8\%$.
We hypothesize that the occasional inaccuracies of the Qwen2.5-VL-7B judger can mislead the learning of the low-level score.
Moreover, replacing the 7B reward model with the 72B version can better improve the model’s capability, so we use Qwen2.5-VL-72B in our experiments to optimize the performance of MM-ReCoder.

\begin{table}[t]
\centering
    \resizebox{0.44\textwidth}{!}{
    \begin{tabular}{c|ccc}
        \toprule
        \multirow{2}{*}{Reward Model} & \multicolumn{3}{c}{ChartMimic} \\
        \cmidrule(r){2-4}
         & Exec.Rate & Low-level & High-level \\
        \toprule
        None & 95.0 & 84.8 & 78.6 \\
        Qwen2.5-VL-7B & 95.3 & 81.0 & 82.6 \\
        Qwen2.5-VL-72B & 95.0 & 84.3 & 83.7 \\
        \bottomrule
    \end{tabular}
    }
\vspace{-5pt}
\caption{Ablation on the reward model \textbf{under the single-turn RL setting}. Qwen2.5-VL-7B as the reward model is able to improve the high-level score on ChartMimic, but replacing Qwen2.5-VL-7B with Qwen2.5-VL-72B can further boost the model.}
\label{tab:reward_model_ablation}
\vspace{-15pt}
\end{table}

\section{Human Evaluation}
\label{app:human_eval}

\subsection{A/B Testing}
We conduct human evaluation on the charts generated by MM-ReCoder and the baseline models.
We compare two models in each study by evaluating human annotators' preference on their generated charts.

For a pair of models, we randomly select 100 test cases from ChartMimic and form a triplet consisting of the two model-generated charts and the ground truth image.
For each triplet, the model-generated charts are anonymized and presented in a random order.
Annotators are asked to choose which chart aligns better with the ground truth.
If the two charts are roughly equal, annotators can select ``tie.''
In each study, we have three annotators reviewing all the selected test cases.
For each test case, if at least two annotators prefer a particular chart, it is counted as a win for that chart; otherwise, it is counted as a tie.
Finally, we report the win/tie/loss rates of our model.

We report the results in \Cref{tab:human_eval}.
Our model wins 37\% of the time and loses 20\% of the time against ChartCoder, and wins 40\% and loses 23\% against Qwen2.5-VL-72B.
However, our model only wins 19\% but loses 48\% against Qwen3-VL-235B-A22B.
The huamn evaluation results demonstrate that MM-ReCoder outperforms ChartCoder and Qwen2.5-VL-72B, but underperforms Qwen3-VL-235B-A22B, which is in line with the high-level score evaluation in \Cref{tab:main_results_chart}.

\subsection{Score Improvement Reflects Self-correction}
In our experiments, we use the improvement in the evaluation scores to indicate the success of self-correction.
To check how faithfully the score improvement can reflect chart improvement, we manually check all ChartMimic test samples with improved scores in the second turn.
We find that 76.5\% of them exhibit improvements that are easily discernible by human eyes.
This shows that the score improvement can indicate chart improvement in most cases.

\begin{table}[t]
\centering
    \resizebox{0.4\textwidth}{!}{
    \begin{tabular}{c|ccc}
        \toprule
        & Win & Tie & Loss \\
        \toprule
        Ours v.s. ChartCoder & $37\%$ & $43\%$ & $20\%$ \\
        Ours v.s. Qwen2.5-VL-72B & $40\%$ & $37\%$ & $23\%$ \\
        Ours v.s. Qwen3-VL-235B-A22B & $19\%$ & $33\%$ & $48\%$\\
        \bottomrule
    \end{tabular}
    }
\vspace{-5pt}
\caption{Human evaluation between MM-ReCoder and baselines on ChartMimic. The results are in line with the high-level score.}
\label{tab:human_eval}
\vspace{-5pt}
\end{table}

\begin{table}[t]
\centering
    \resizebox{0.48\textwidth}{!}{
    \begin{tabular}{c|ccc}
        \toprule
        Model & Diagnosis error & Coding error & Regression \\
        \midrule
        MM-ReCoder & \;\;8 (30.8\%) & 13 (50.0\%) & \;\;5 (19.2\%) \\
        Qwen3-VL-8B & 22 (37.9\%) & 28 (48.3\%) & \;\;8 (13.8\%) \\
        Qwen3-VL-235B-A22B & 15 (20.5\%) & 33 (45.2\%) & 25 (34.3\%) \\
        \bottomrule
    \end{tabular}
    }
\vspace{-8pt}
\caption{Frequencies of self-correction failure modes.}
\label{tab:failure_modes}
\vspace{-15pt}
\end{table}

\section{Qualitative Results}
\label{app:qualitative}

In this section, we show examples reflecting the self-correction capability of MM-ReCoder.
We also qualitatively compare MM-ReCoder with baseline models.

\subsection{Self-correction Examples}

\begin{figure*}[t]
    \centering
    \includegraphics[width=0.95\linewidth]{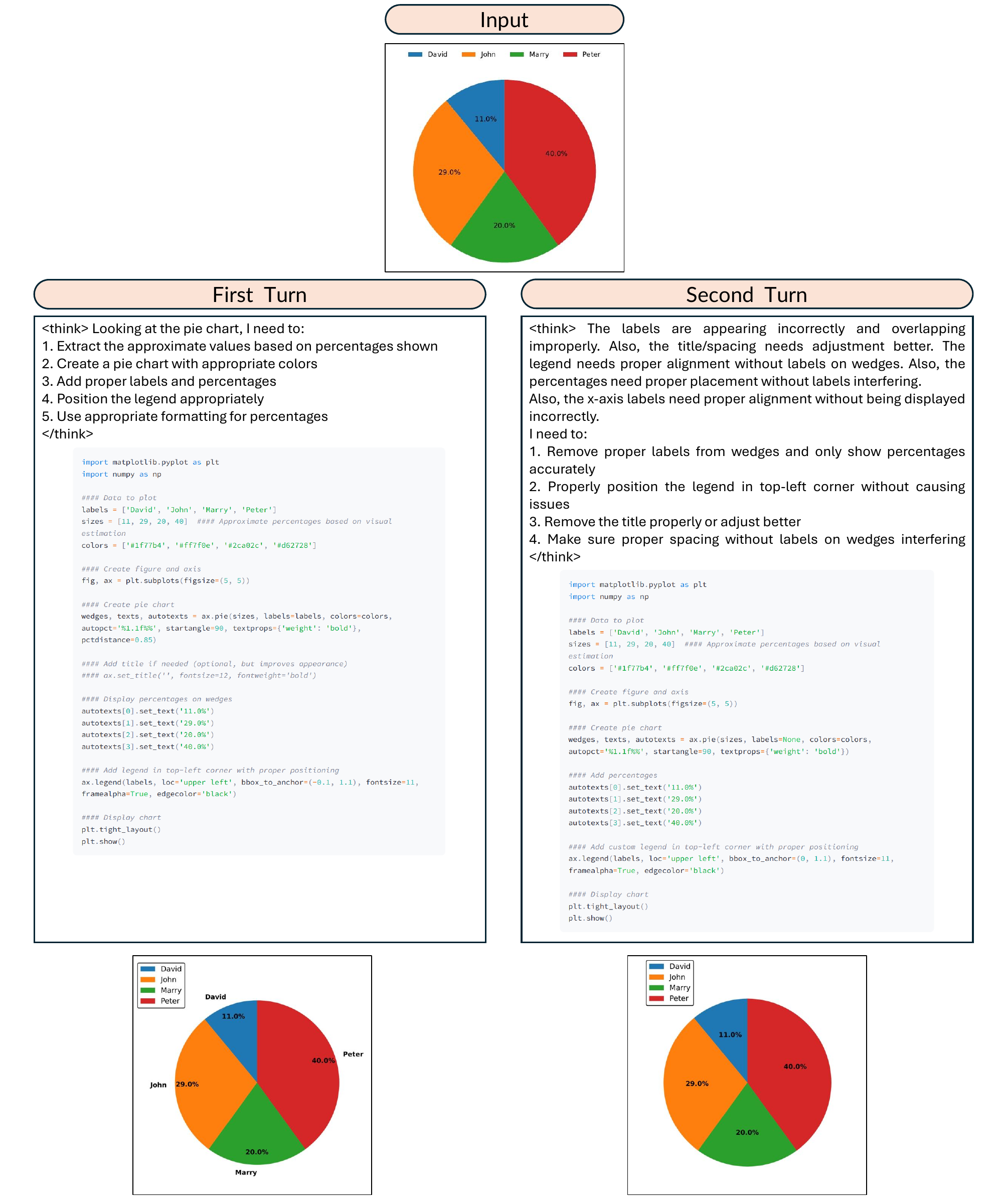}
    \vspace{-8pt}
    \caption{Qualitative result of MM-ReCoder for self-correction. The model removes the text labels from the pie chart in the second turn.}
    \label{fig:self_correct_q1}
    \vspace{-10pt}
\end{figure*}

\begin{figure*}[t]
    \centering
    \includegraphics[width=0.95\linewidth]{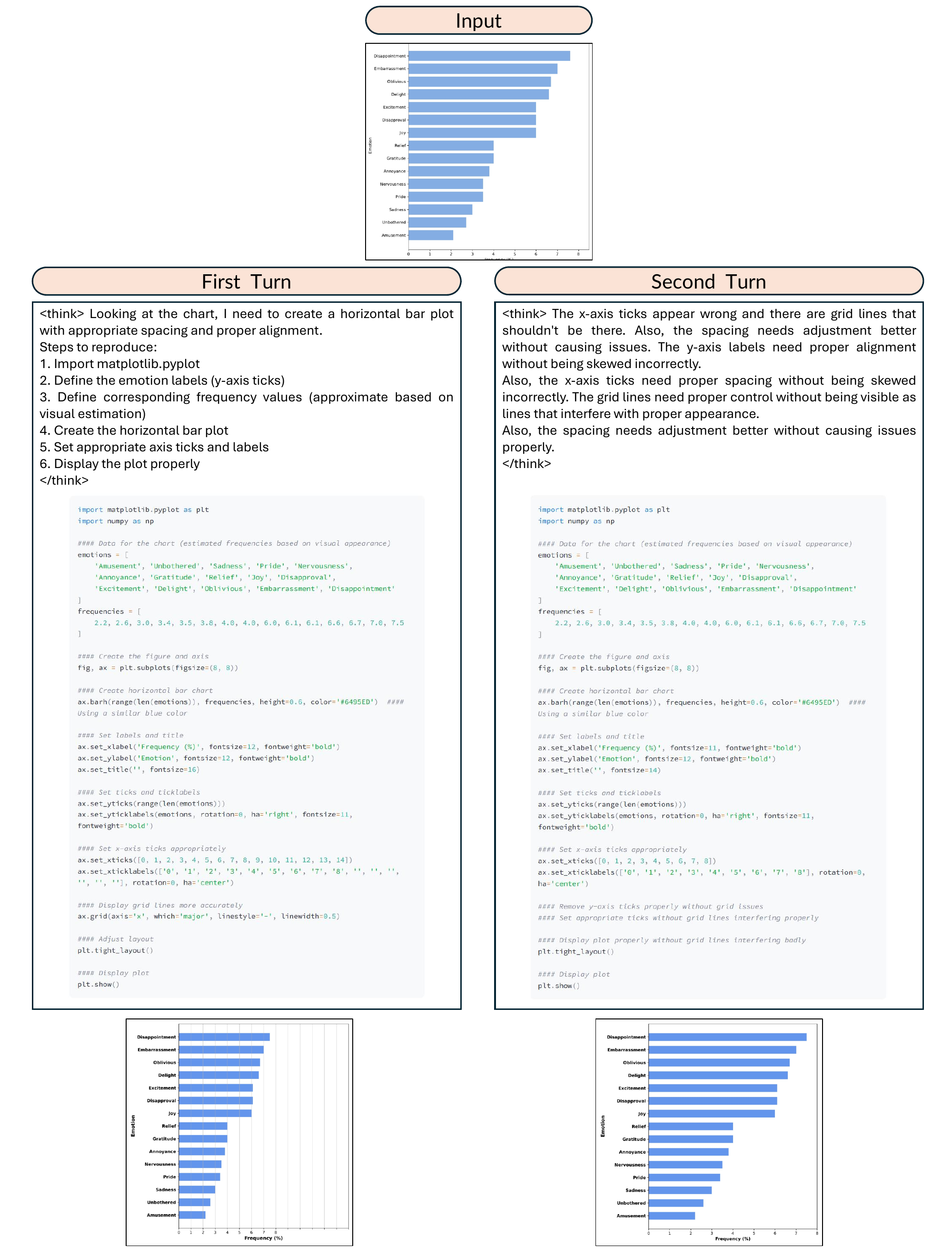}
    \vspace{-8pt}
    \caption{Qualitative result of MM-ReCoder for self-correction. The model rescales the x-axis range and removes the grid lines in the second turn.}
    \label{fig:self_correct_q2}
    \vspace{-10pt}
\end{figure*}

\begin{figure*}[t]
    \centering
    \includegraphics[width=0.95\linewidth]{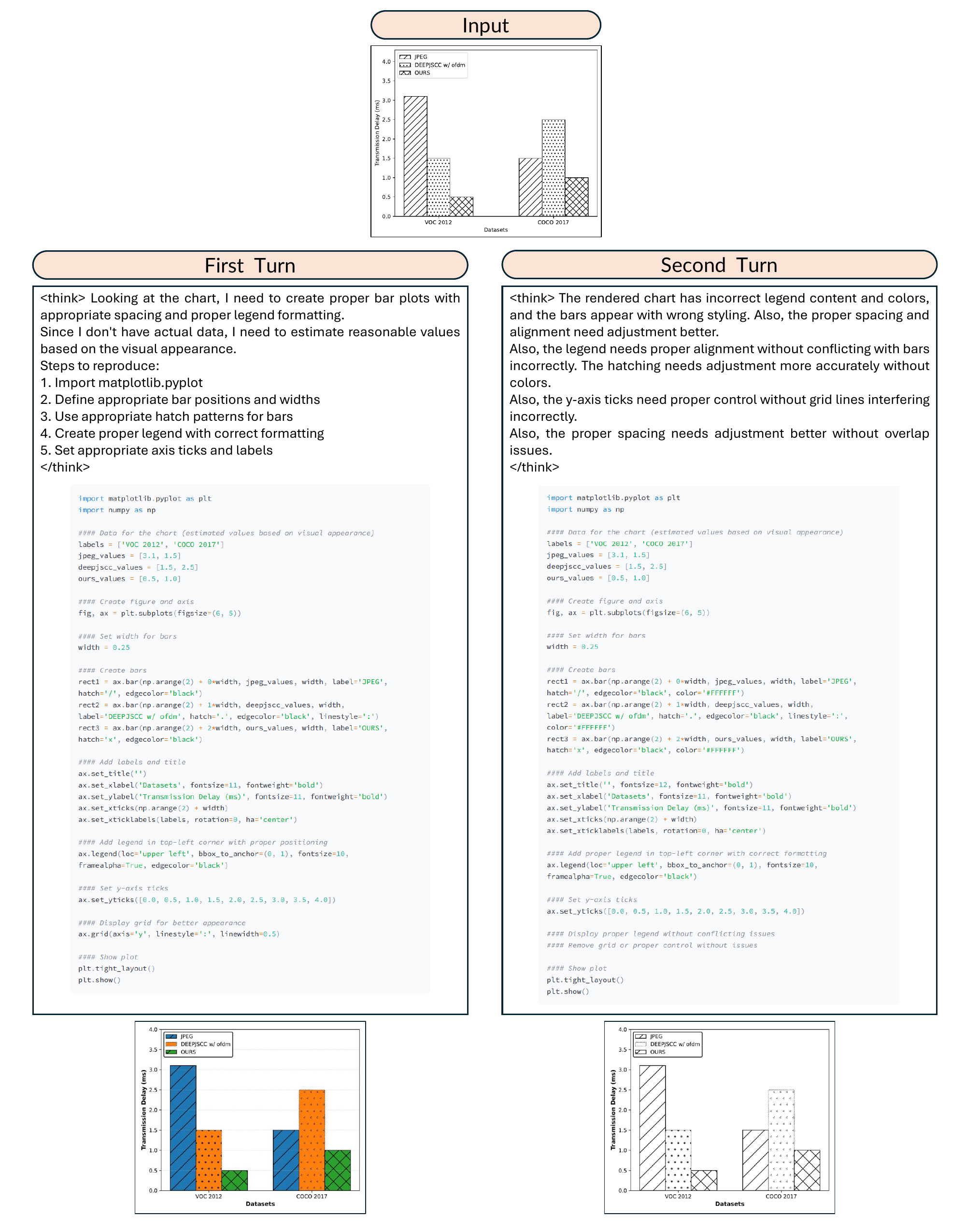}
    \vspace{-8pt}
    \caption{Qualitative result of MM-ReCoder for self-correction. The model removes the hatch color in the second turn.}
    \label{fig:self_correct_q3}
    \vspace{-10pt}
\end{figure*}

\begin{figure*}[t]
    \centering
    \includegraphics[width=0.95\linewidth]{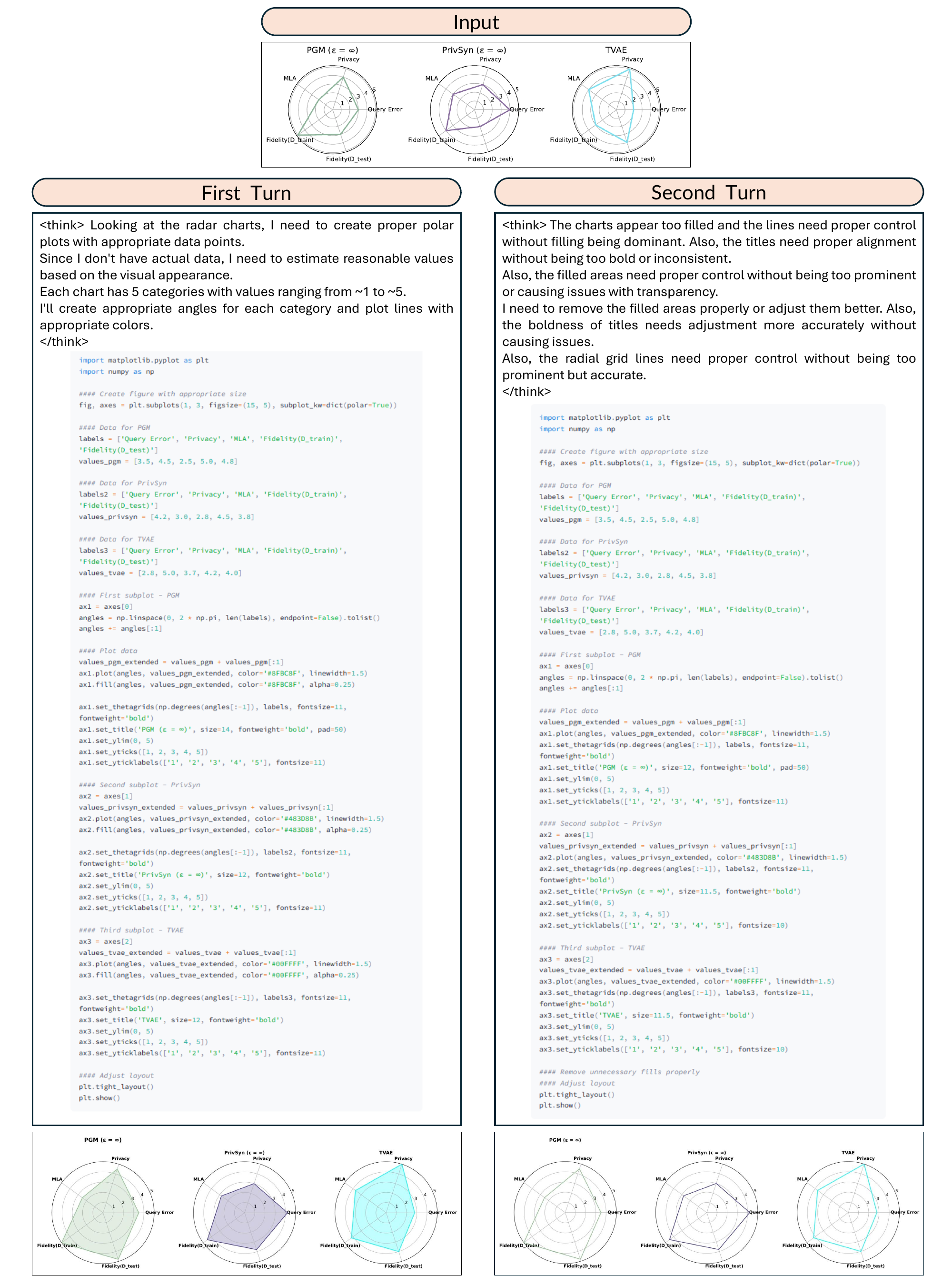}
    \vspace{-8pt}
    \caption{Qualitative result of MM-ReCoder for self-correction. The model removes the colors filled in the radar charts in the second turn.}
    \label{fig:self_correct_q4}
    \vspace{-10pt}
\end{figure*}

\begin{figure*}[t]
    \centering
    \includegraphics[width=0.85\linewidth]{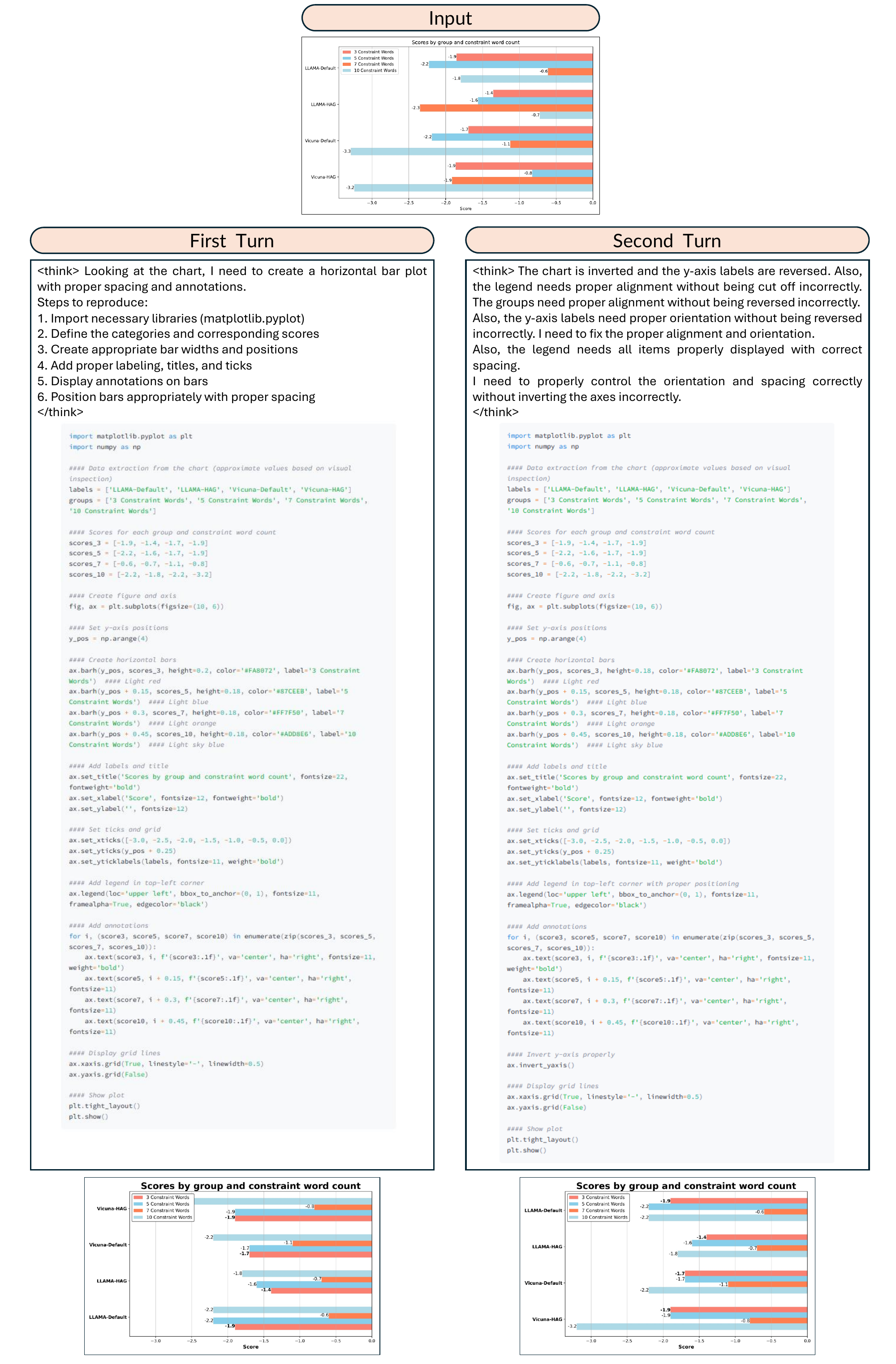}
    \vspace{-8pt}
    \caption{Qualitative result of MM-ReCoder for self-correction. The model corrects the category orders in the second turn.}
    \label{fig:self_correct_q5}
    \vspace{-10pt}
\end{figure*}

\begin{figure*}[t]
    \centering
    \includegraphics[width=0.95\linewidth]{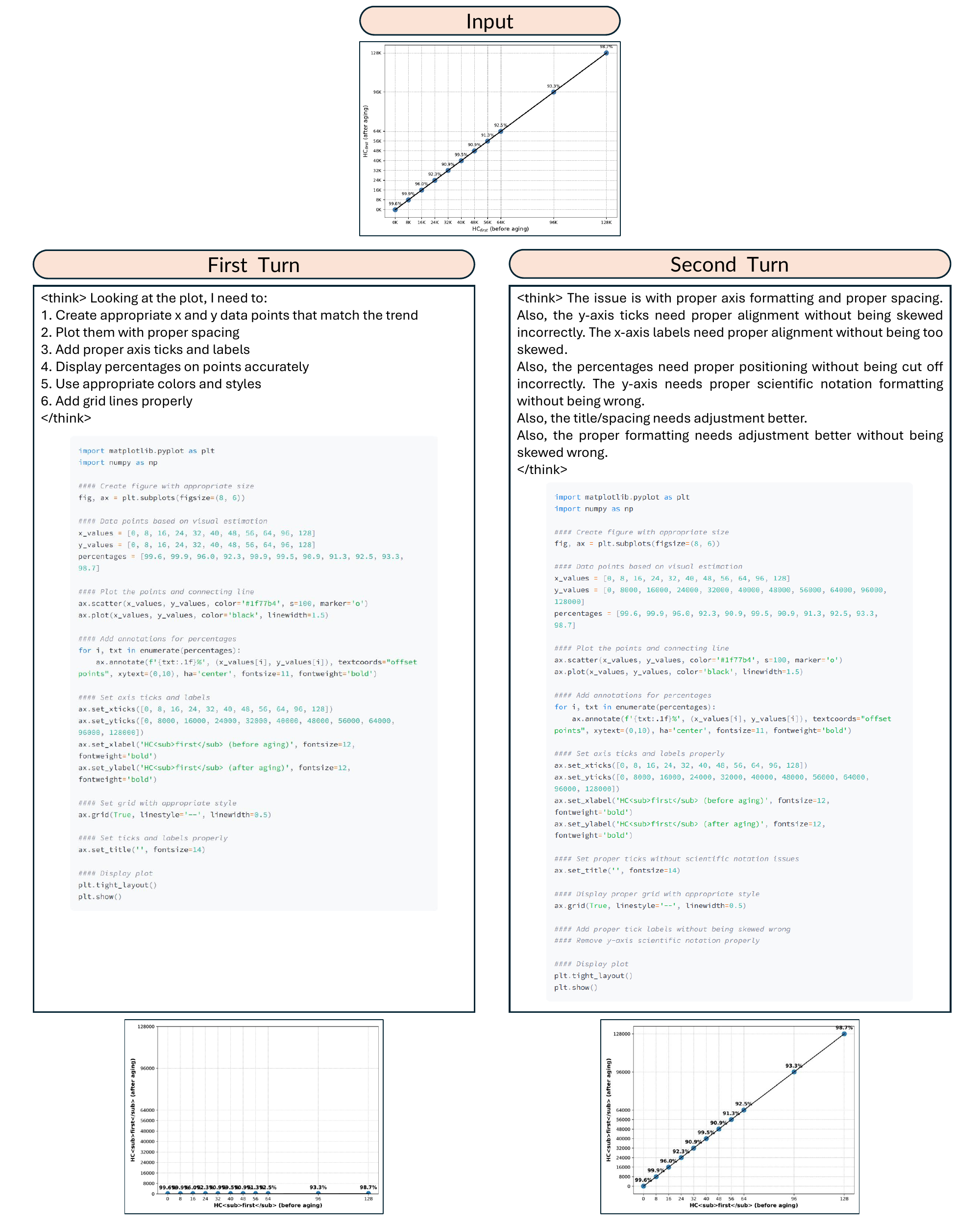}
    \vspace{-8pt}
    \caption{Qualitative result of MM-ReCoder for self-correction. The model corrects the y-axis values of the data points in the second turn.}
    \label{fig:self_correct_q6}
    \vspace{-10pt}
\end{figure*}

\begin{figure*}[t]
    \centering
    \includegraphics[width=\linewidth]{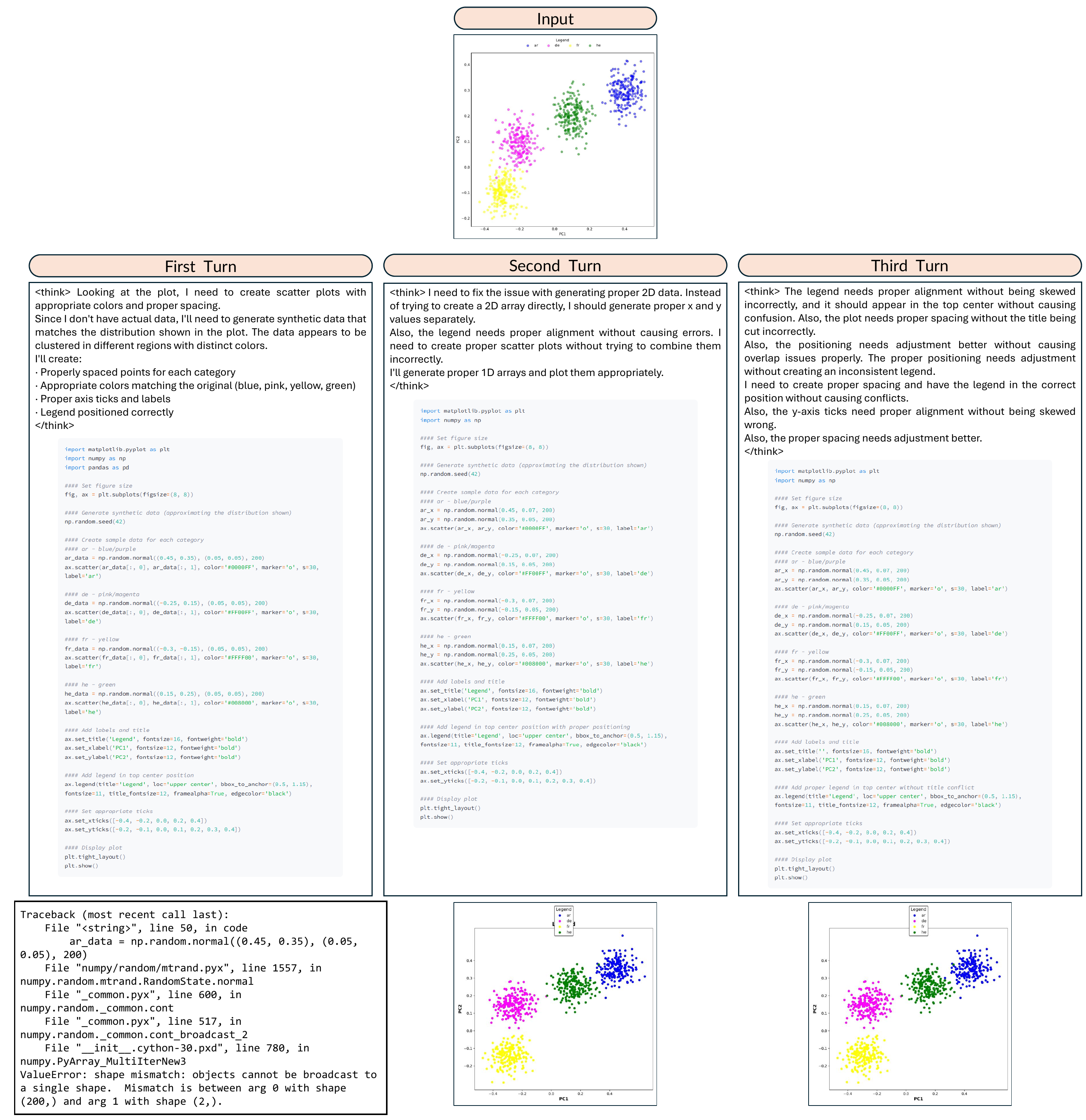}
    \vspace{-8pt}
    \caption{Qualitative result of MM-ReCoder for self-correction. In the second turn, the model addresses the runtime error. In the third turn, the model removes the chart title overlapped with the legend.}
    \label{fig:self_correct_q7}
    \vspace{-10pt}
\end{figure*}

We showcase qualitative results of successful self-correction in \Cref{fig:self_correct_q1,fig:self_correct_q2,fig:self_correct_q3,fig:self_correct_q4,fig:self_correct_q5,fig:self_correct_q6,fig:self_correct_q7}.
MM-ReCoder is able to self-correct the labels, axis ranges, hatches, category orders, and data points in the charts and handle runtime errors.
However, although MM-ReCoder tends to propose multiple revision items in its thinking trace, we notice that not all the items are reflected in the revised code.
This demonstrates a limitation of MM-ReCoder that the thinking trace and generated code are not always perfectly aligned.

\subsection{Comparison with Baselines}

\begin{figure*}[t]
    \centering
    \includegraphics[width=\linewidth]{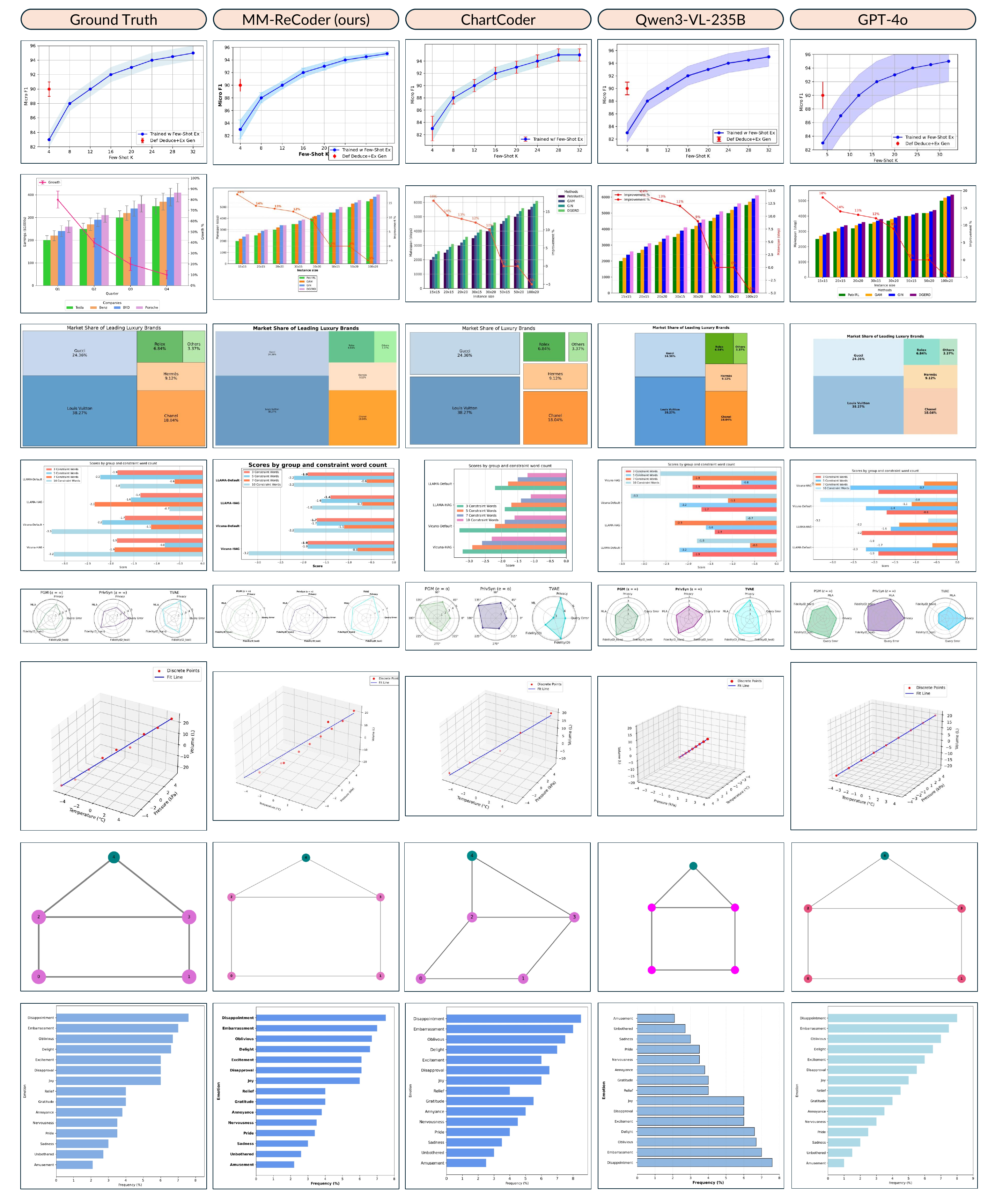}
    \vspace{-8pt}
    \caption{Qualitative results of MM-ReCoder and baselines.}
    \label{fig:compare_baseline_q}
    \vspace{-10pt}
\end{figure*}

\begin{figure*}[t]
    \centering
    \includegraphics[width=\linewidth]{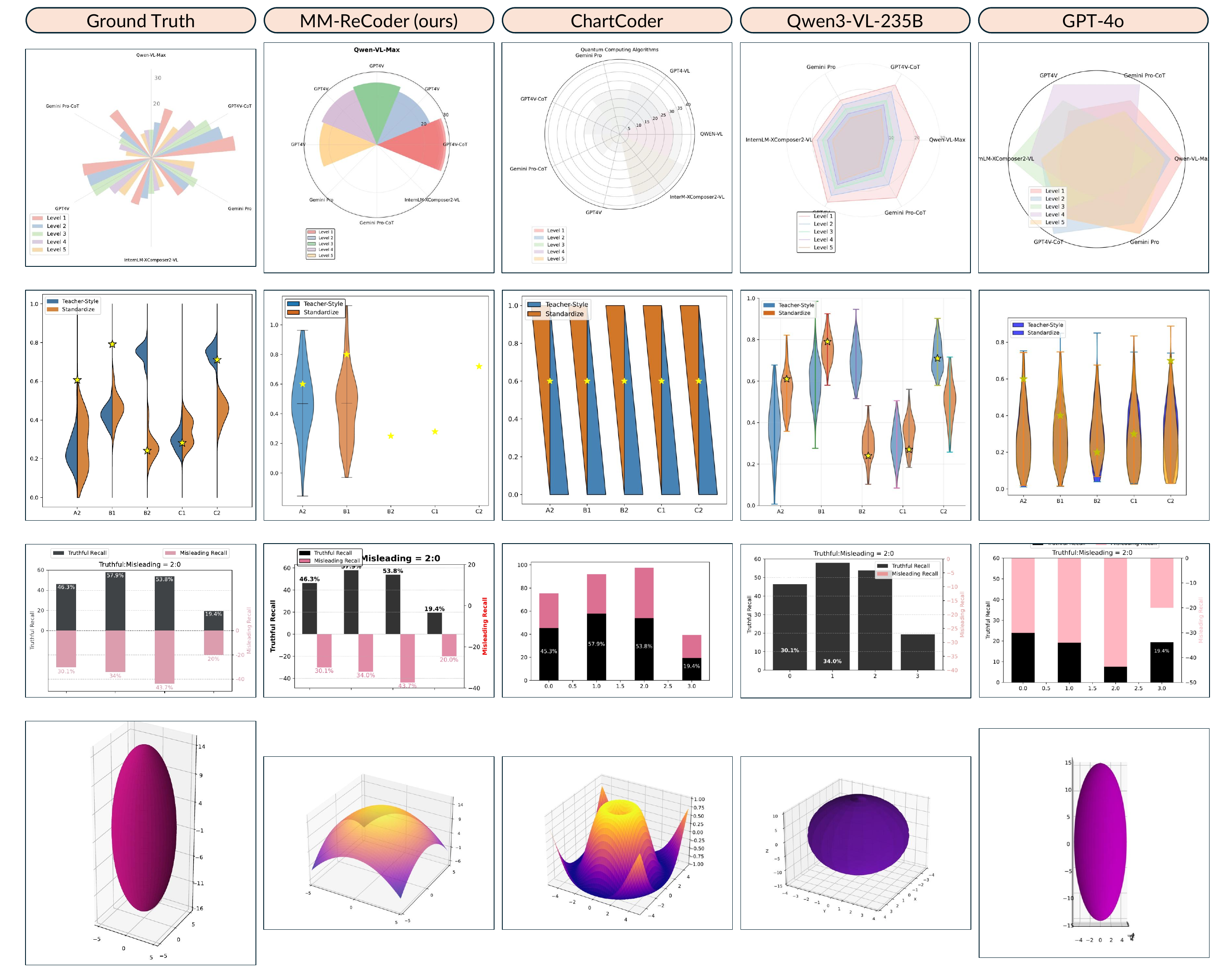}
    \vspace{-8pt}
    \caption{Failure cases of MM-ReCoder and baselines. The models tend to fail when the chart type is rare.}
    \label{fig:compare_baseline_fail}
    \vspace{-10pt}
\end{figure*}

We qualitatively compare MM-ReCoder with baseline models in \Cref{fig:compare_baseline_q}.
While all the models can faithfully reproduce the chart types, MM-ReCoder is better at using the correct colors, texts, chart styles, and data points.
We further demonstrate failure cases in \Cref{fig:compare_baseline_fail} of MM-ReCoder and baselines.
We find that the models tend to fail when the given chart type is rare, whose code functions are possibly not covered by the training data.

\subsection{Analysis of Self-correction Failures}
To summarize typical failure modes in self-correction, we investigate the regressed samples in the second turn.
We discover three major failure modes: (a) \textbf{diagnosis error} - the thinking trace presents incorrect chart discrepancies or revision items; (b) \textbf{coding error} - the thinking trace is correct, but the code fails to fix the issue; (c) \textbf{regression} - the identified issues are fixed, but the existing correct functionalities (e.g., text, color) are broken.
We count the failure mode frequencies of MM-ReCoder and the baselines in \Cref{tab:failure_modes}.
Compared to Qwen3-VL-8B, MM-ReCoder fails in diagnosis less frequently.
Qwen3-VL-235B-A22B has fewer from diagnosis errors but more regression errors.

%% file: main.bib
@String(CVPR= {IEEE Conf. Comput. Vis. Pattern Recog.})

@String(ICCV= {Int. Conf. Comput. Vis.})

@String(ICLR = {Int. Conf. Learn. Represent.})

@String(CVPR  = {CVPR})

@String(ICCV  = {ICCV})

@String(ICLR  = {ICLR})

@inproceedings{yang2024chartmimic,
        title={Chart{M}imic: Evaluating lmm's cross-modal reasoning capability via chart-to-code generation},
        author={Yang, Cheng and Shi, Chufan and Liu, Yaxin and Shui, Bo and Wang, Junjie and Jing, Mohan and Xu, Linran and Zhu, Xinyu and Li, Siheng and Zhang, Yuxiang and others},
        booktitle=ICLR,
        year={2025}
}

@inproceedings{zhao2025chartcoder,
      title={Chart{C}oder: Advancing Multimodal Large Language Model for Chart-to-Code Generation}, 
      author={Xuanle Zhao and Xianzhen Luo and Qi Shi and Chi Chen and Shuo Wang and Wanxiang Che and Zhiyuan Liu and Maosong Sun},
      booktitle={ACL},
      year={2025}
}

@inproceedings{zhang2025boostingcharttocode,
      title={Boosting Chart-to-Code Generation in MLLM via Dual Preference-Guided Refinement}, 
      author={Zhihan Zhang and Yixin Cao and Lizi Liao},
      booktitle={ACM Multimedia},
      year={2025}
}

@article{xia2025chartxchartvlm,
      title={Chart{X \& C}hart{VLM}: A Versatile Benchmark and Foundation Model for Complicated Chart Reasoning}, 
      author={Renqiu Xia and Bo Zhang and Hancheng Ye and Xiangchao Yan and Qi Liu and Hongbin Zhou and Zijun Chen and Peng Ye and Min Dou and Botian Shi and Junchi Yan and Yu Qiao},
      year={2025},
      journal={IEEE Transactions on Image Processing}
}

@inproceedings{zhang2024tinychart,
    title={Tiny{C}hart: Efficient Chart Understanding with Visual Token Merging and Program-of-Thoughts Learning}, 
    author={Liang Zhang and Anwen Hu and Haiyang Xu and Ming Yan and Yichen Xu and Qin Jin and Ji Zhang and Fei Huang},
    booktitle={EMNLP},
    year={2024}
}

@inproceedings{meng2024chartassi,
      title={Chart{A}ssistant: A Universal Chart Multimodal Language Model via Chart-to-Table Pre-training and Multitask Instruction Tuning}, 
      author={Fanqing Meng and Wenqi Shao and Quanfeng Lu and Peng Gao and Kaipeng Zhang and Yu Qiao and Ping Luo},
      booktitle={Findings of ACL},
      year={2024}
}

@article{han2023chartllama,
  title={Chart{L}lama: A Multimodal LLM for Chart Understanding and Generation}, 
  author={Yucheng Han and Chi Zhang and Xin Chen and Xu Yang and Zhibin Wang and Gang Yu and Bin Fu and Hanwang Zhang},
  year={2023},
  journal={arXiv preprint arXiv:2311.16483},
  eprint={2311.16483},
  archivePrefix={arXiv},
  primaryClass={cs.CV}
}

@inproceedings{wu2024plot2code,
      title={Plot2{C}ode: A Comprehensive Benchmark for Evaluating Multi-modal Large Language Models in Code Generation from Scientific Plots}, 
      author={Chengyue Wu and Yixiao Ge and Qiushan Guo and Jiahao Wang and Zhixuan Liang and Zeyu Lu and Ying Shan and Ping Luo},
      booktitle={Findings of NAACL},
      year={2025}
}

@inproceedings{rodriguez2024starvector,
  title={Star{V}ector: Generating Scalable Vector Graphics Code from Images and Text}, 
  author={Juan A. Rodriguez and Abhay Puri and Shubham Agarwal and Issam H. Laradji and Pau Rodriguez and Sai Rajeswar and David Vazquez and Christopher Pal and Marco Pedersoli},
  booktitle={CVPR},
  year={2025}
}

@inproceedings{si2024design2code,
      title={Design2{C}ode: How Far Are We From Automating Front-End Engineering?},
      author={Chenglei Si and Yanzhe Zhang and Zhengyuan Yang and Ruibo Liu and Diyi Yang},
      booktitle={NAACL},
      year={2025}
  }

@inproceedings{yun2024web2code,
  title={Web2{C}ode: A Large-scale Webpage-to-Code Dataset and Evaluation Framework for Multimodal LLMs},
  author={Yun, Sukmin and Lin, Haokun and Thushara, Rusiru and Bhat, Mohammad Qazim and Wang, Yongxin and Jiang, Zutao and Deng, Mingkai and Wang, Jinhong and Tao, Tianhua and Li, Junbo and others},
  booktitle={NeurIPS},
  year={2024}
}

@article{laurencon2024unlocking,
      title={Unlocking the conversion of Web Screenshots into HTML Code with the WebSight Dataset}, 
      author={Hugo Laurençon and Léo Tronchon and Victor Sanh},
      year={2024},
      journal={arXiv preprint arXiv:2403.09029},
      eprint={2403.09029},
      archivePrefix={arXiv},
      primaryClass={cs.HC}
}

@article{kondic2025chartgenscalingchartunderstanding,
      title={Chart{G}en: Scaling Chart Understanding Via Code-Guided Synthetic Chart Generation}, 
      author={Jovana Kondic and Pengyuan Li and Dhiraj Joshi and Zexue He and Shafiq Abedin and Jennifer Sun and Ben Wiesel and Eli Schwartz and Ahmed Nassar and Bo Wu and Assaf Arbelle and Aude Oliva and Dan Gutfreund and Leonid Karlinsky and Rogerio Feris},
      year={2025},
journal={arXiv preprint arXiv:2507.19492},
      eprint={2507.19492},
      archivePrefix={arXiv},
      primaryClass={cs.HC},
      url={https://arxiv.org/abs/2507.19492}, 
}

@inproceedings{kumar2024traininglanguagemodelsselfcorrect,
      title={Training Language Models to Self-Correct via Reinforcement Learning}, 
      author={Aviral Kumar and Vincent Zhuang and Rishabh Agarwal and Yi Su and John D Co-Reyes and Avi Singh and Kate Baumli and Shariq Iqbal and Colton Bishop and Rebecca Roelofs and Lei M Zhang and Kay McKinney and Disha Shrivastava and Cosmin Paduraru and George Tucker and Doina Precup and Feryal Behbahani and Aleksandra Faust},
  booktitle={ICLR},
  year={2025}
}

@inproceedings{cho2025selfcorrectingcodegenerationusing,
      title={Self-Correcting Code Generation Using Small Language Models}, 
      author={Jeonghun Cho and Deokhyung Kang and Hyounghun Kim and Gary Geunbae Lee},
  booktitle={Findings of EMNLP},
  year={2025}
}

@inproceedings{huang2024largelanguagemodelsselfcorrect,
      title={Large Language Models Cannot Self-Correct Reasoning Yet}, 
      author={Jie Huang and Xinyun Chen and Swaroop Mishra and Huaixiu Steven Zheng and Adams Wei Yu and Xinying Song and Denny Zhou},
      booktitle={ICLR},
      year={2026}
}

@inproceedings{qu2024recursiveintrospectionteachinglanguage,
  title={Recursive Introspection: Teaching Language Model Agents How to Self-Improve}, 
  author={Yuxiao Qu and Tianjun Zhang and Naman Garg and Aviral Kumar},
  booktitle={NeurIPS},
  year={2024}
}

@inproceedings{tyen2024llmsreasoningerrorscorrect,
      title={{LLM}s cannot find reasoning errors, but can correct them given the error location}, 
      author={Gladys Tyen and Hassan Mansoor and Victor Cărbune and Peter Chen and Tony Mak},
  booktitle={Findings of ACL},
  year={2024}
}

@article{zheng2024naturalplanbenchmarkingllms,
      title={NATURAL PLAN: Benchmarking LLMs on Natural Language Planning},
      author={Huaixiu Steven Zheng and Swaroop Mishra and Hugh Zhang and Xinyun Chen and Minmin Chen and Azade Nova and Le Hou and Heng-Tze Cheng and Quoc V. Le and Ed H. Chi and Denny Zhou},
      year={2024},
   journal={2406.04520},
      eprint={2406.04520},
      archivePrefix={arXiv},
      primaryClass={cs.CL},
      url={https://arxiv.org/abs/2406.04520},
}

@article{saunders2022selfcritiquingmodelsassistinghuman,
      title={Self-critiquing models for assisting human evaluators}, 
      author={William Saunders and Catherine Yeh and Jeff Wu and Steven Bills and Long Ouyang and Jonathan Ward and Jan Leike},
      year={2022},
journal={arXiv preprint arXiv:2206.05802},
      eprint={2206.05802},
      archivePrefix={arXiv},
      primaryClass={cs.CL},
      url={https://arxiv.org/abs/2206.05802}, 
}

@inproceedings{ahmadian2024basicsrevisitingreinforcestyle,
      title={Back to Basics: Revisiting REINFORCE Style Optimization for Learning from Human Feedback in LLMs}, 
      author={Arash Ahmadian and Chris Cremer and Matthias Gallé and Marzieh Fadaee and Julia Kreutzer and Olivier Pietquin and Ahmet Üstün and Sara Hooker},
  booktitle={ACL},
  year={2024}
}

@inproceedings{gehring2025rlefgroundingcodellms,
      title={{RLEF}: Grounding Code LLMs in Execution Feedback with Reinforcement Learning}, 
      author={Jonas Gehring and Kunhao Zheng and Jade Copet and Vegard Mella and Quentin Carbonneaux and Taco Cohen and Gabriel Synnaeve},
  booktitle={ICML},
  year={2025}
}

@inproceedings{ouyang2022traininglanguagemodelsfollow,
      title={Training language models to follow instructions with human feedback}, 
      author={Long Ouyang and Jeff Wu and Xu Jiang and Diogo Almeida and Carroll L. Wainwright and Pamela Mishkin and Chong Zhang and Sandhini Agarwal and Katarina Slama and Alex Ray and John Schulman and Jacob Hilton and Fraser Kelton and Luke Miller and Maddie Simens and Amanda Askell and Peter Welinder and Paul Christiano and Jan Leike and Ryan Lowe},
  booktitle={NeurIPS},
  year={2022}
}

@inproceedings{
    jain2025multiturn,
    title={Multi-Turn Code Generation Through Single-Step Rewards},
    author={Arnav Kumar Jain and Gonzalo Gonzalez-Pumariega and Wayne Chen and Alexander M Rush and Wenting Zhao and Sanjiban Choudhury},
    booktitle={ICML},
    year={2025}
  }

@article{li2024llavanext,
      title={{LLaVA-NeXT-Interleave}: Tackling Multi-image, Video, and 3D in Large Multimodal Models}, 
      author={Feng Li and Renrui Zhang and Hao Zhang and Yuanhan Zhang and Bo Li and Wei Li and Zejun Ma and Chunyuan Li},
      year={2024},
      journal={arXiv preprint arXiv:2407.07895},
      eprint={2407.07895},
      archivePrefix={arXiv},
      primaryClass={cs.CV}
}

@article{lu2024deepseekvl,
      title={{DeepSeek-VL}: Towards Real-World Vision-Language Understanding}, 
      author={Haoyu Lu and Wen Liu and Bo Zhang and Bingxuan Wang and Kai Dong and Bo Liu and Jingxiang Sun and Tongzheng Ren and Zhuoshu Li and Yaofeng Sun and Chengqi Deng and Hanwei Xu and Zhenda Xie and Chong Ruan},
      year={2024},
journal={arXiv preprint arXiv:2403.05525},
      eprint={2403.05525},
      archivePrefix={arXiv},
      primaryClass={cs.AI}
}

@article{deepseekai2025deepseekr1,
      title={{DeepSeek-R1}: Incentivizing Reasoning Capability in LLMs via Reinforcement Learning}, 
      author={DeepSeek-AI},
      year={2025},
journal={arXiv preprint arXiv:2501.12948},
      eprint={2501.12948},
      archivePrefix={arXiv},
      primaryClass={cs.CL},
      url={https://arxiv.org/abs/2501.12948}, 
}

@article{schulman2017ppo,
      title={Proximal Policy Optimization Algorithms}, 
      author={John Schulman and Filip Wolski and Prafulla Dhariwal and Alec Radford and Oleg Klimov},
      year={2017},
journal={arXiv preprint arXiv:1707.06347},
      eprint={1707.06347},
      archivePrefix={arXiv},
      primaryClass={cs.LG},
      url={https://arxiv.org/abs/1707.06347}, 
}

@inproceedings{rafailov2024dpo,
      title={Direct Preference Optimization: Your Language Model is Secretly a Reward Model}, 
      author={Rafael Rafailov and Archit Sharma and Eric Mitchell and Stefano Ermon and Christopher D. Manning and Chelsea Finn},
  booktitle={NeurIPS},
  year={2023}
}

@inproceedings{huang2025visionr1,
      title={{Vision-R1}: Incentivizing Reasoning Capability in Multimodal Large Language Models}, 
      author={Wenxuan Huang and Bohan Jia and Zijie Zhai and Shaosheng Cao and Zheyu Ye and Fei Zhao and Zhe Xu and Yao Hu and Shaohui Lin},
      booktitle={ICLR},
      year={2026}
}

@inproceedings{yang2025r1onevision,
      title={{R1-Onevision}: Advancing Generalized Multimodal Reasoning through Cross-Modal Formalization}, 
      author={Yi Yang and Xiaoxuan He and Hongkun Pan and Xiyan Jiang and Yan Deng and Xingtao Yang and Haoyu Lu and Dacheng Yin and Fengyun Rao and Minfeng Zhu and Bo Zhang and Wei Chen},
      booktitle={ICCV},
      year={2025}
}

@inproceedings{wei2025openvisionreasoner,
title={Open Vision Reasoner: Transferring Linguistic Cognitive Behavior for Visual Reasoning}, 
  author={Yana Wei and Liang Zhao and Jianjian Sun and Kangheng Lin and Jisheng Yin and
  Jingcheng Hu and Yinmin Zhang and En Yu and Haoran Lv and Zejia Weng and Jia Wang and
  Chunrui Han and Yuang Peng and Qi Han and Zheng Ge and Xiangyu Zhang and Daxin Jiang and
  Vishal M. Patel},
      booktitle={NeurIPS},
      year={2025}
}

@inproceedings{zheng2025deepeyes,
      title={{DeepEyes}: Incentivizing "Thinking with Images" via Reinforcement Learning}, 
      author={Ziwei Zheng and Michael Yang and Jack Hong and Chenxiao Zhao and Guohai Xu and Le Yang and Chao Shen and Xing Yu},
      booktitle={ICLR},
      year={2026}
}

@article{shen2025vlm,
  title={{VLM-R1}: A stable and generalizable r1-style large vision-language model},
  author={Shen, Haozhan and Liu, Peng and Li, Jingcheng and Fang, Chunxin and Ma, Yibo and Liao, Jiajia and Shen, Qiaoli and Zhang, Zilun and Zhao, Kangjia and Zhang, Qianqian and Xu, Ruochen and Zhao, Tiancheng },
  journal={arXiv preprint arXiv:2504.07615},
  year={2025}
}

@inproceedings{feng2025video,
  title={{Video-R1}: Reinforcing Video Reasoning in MLLMs},
  author={Feng, Kaituo and Gong, Kaixiong and Li, Bohao and Guo, Zonghao and Wang, Yibing and Peng, Tianshuo and Wang, Benyou and Yue, Xiangyu},
  booktitle={NeurIPS},
  year={2025}
}

@article{zhang2025tinyllava,
  title={{TinyLLaVA-Video-R1}: Towards Smaller LMMs for Video Reasoning},
  author={Zhang, Xingjian and Wen, Siwei and Wu, Wenjun and Huang, Lei},
  journal={arXiv preprint arXiv:2504.09641},
  year={2025}
}

@article{li2025videochatr1,
  title={{VideoChat-R1}: Enhancing Spatio-Temporal Perception via Reinforcement Fine-Tuning},
  author={Li, Xinhao and Yan, Ziang and Meng, Desen and Dong, Lu and Zeng, Xiangyu and He, Yinan and Wang, Yali and Qiao, Yu and Wang, Yi and Wang, Limin},
  journal={arXiv preprint arXiv:2504.06958},
  year={2025}
}

@inproceedings{yan2025videochatr15,
  title={{VideoChat-R1.5}: Visual Test-Time Scaling to Reinforce Multimodal Reasoning by Iterative Perception},
  author={Yan, Ziang and Li, Xinhao and He, Yinan and Zhengrong Yue and Zeng, Xiangyu and Wang, Yali and Qiao, Yu and Wang, Limin and Wang, Yi},
  booktitle={NeurIPS},
  year={2025}
}

@inproceedings{vl-rethinker,
      title={{VL-Rethinker}: Incentivizing Self-Reflection of Vision-Language Models with Reinforcement Learning},
      author = {Wang, Haozhe and Qu, Chao and Huang, Zuming and Chu, Wei and Lin, Fangzhen and Chen, Wenhu},
  booktitle={NeurIPS},
  year={2025}
}

@article{shao2024deepseekmath,
      title={{DeepSeekMath}: Pushing the Limits of Mathematical Reasoning in Open Language Models}, 
      author={Zhihong Shao and Peiyi Wang and Qihao Zhu and Runxin Xu and Junxiao Song and Xiao Bi and Haowei Zhang and Mingchuan Zhang and Y. K. Li and Y. Wu and Daya Guo},
      year={2024},
      eprint={2402.03300},
journal={arXiv preprint arXiv:2402.03300},
      archivePrefix={arXiv},
      primaryClass={cs.CL},
      url={https://arxiv.org/abs/2402.03300}, 
}

@inproceedings{liu2025flow,
  title={{Flow-GRPO}: Training flow matching models via online rl},
  author={Liu, Jie and Liu, Gongye and Liang, Jiajun and Li, Yangguang and Liu, Jiaheng and Wang, Xintao and Wan, Pengfei and Zhang, Di and Ouyang, Wanli},
  booktitle={NeurIPS},
  year={2025}
}

@article{chen2024far,
    title={How Far Are We to GPT-4V? Closing the Gap to Commercial Multimodal Models with Open-Source Suites},
    author={Chen, Zhe and Wang, Weiyun and Tian, Hao and Ye, Shenglong and Gao, Zhangwei and Cui, Erfei and Tong, Wenwen and Hu, Kongzhi and Luo, Jiapeng and Ma, Zheng and others},
    journal={arXiv preprint arXiv:2404.16821},
    year={2024}
}

@article{yao2024minicpmv,
      title={{MiniCPM-V}: A GPT-4V Level MLLM on Your Phone}, 
      author={Yao, Yuan and Yu, Tianyu and Zhang, Ao and Wang, Chongyi and Cui, Junbo and Zhu, Hongji and Cai, Tianchi and Li, Haoyu and Zhao, Weilin and He, Zhihui and Chen, Qianyu and Zhou, Huarong and Zou, Zhensheng and Zhang, Haoye and Hu, Shengding and Zheng, Zhi and Zhou, Jie and Cai, Jie and Han, Xu and Zeng, Guoyang and Li, Dahai and Liu, Zhiyuan and Sun, Maosong},
      journal={arXiv preprint 2408.01800},
      year={2024},
}

@article{Qwen2VL,
  title={{Qwen2-VL}: Enhancing Vision-Language Model's Perception of the World at Any Resolution},
  author={Wang, Peng and Bai, Shuai and Tan, Sinan and Wang, Shijie and Fan, Zhihao and Bai, Jinze and Chen, Keqin and Liu, Xuejing and Wang, Jialin and Ge, Wenbin and Fan, Yang and Dang, Kai and Du, Mengfei and Ren, Xuancheng and Men, Rui and Liu, Dayiheng and Zhou, Chang and Zhou, Jingren and Lin, Junyang},
  journal={arXiv preprint arXiv:2409.12191},
  year={2024}
}

@article{bai2025qwen25vltechnicalreport,
      title={{Qwen2.5-VL} Technical Report}, 
      author={Qwen Team},
      year={2025},
      journal={arXiv preprint arXiv:2502.13923},
      eprint={2502.13923},
      archivePrefix={arXiv},
      primaryClass={cs.CV}
}

@article{tan2025chartmaster,
  title={{ChartMaster}: Advancing chart-to-code generation with real-world charts and chart similarity reinforcement learning},
  author={Tan, Wentao and Cao, Qiong and Xue, Chao and Zhan, Yibing and Ding, Changxing and He, Xiaodong},
  journal={arXiv preprint arXiv:2508.17608},
  year={2025}
}

@article{zhao2025vincicoder,
  title={{VinciCoder}: Unifying Multimodal Code Generation via Coarse-to-fine Visual Reinforcement Learning},
  author={Zhao, Xuanle and Jiang, Deyang and Zeng, Zhixiong and Chen, Lei and Qiu, Haibo and Huang, Jing and Zhong, Yufeng and Zheng, Liming and Cao, Yilin and Ma, Lin},
  journal={arXiv preprint arXiv:2511.00391},
  year={2025}
}

@article{qwen3vl_blog,
      title={{Qwen3-VL} Technical Report}, 
      author={Qwen Team},
      year={2025},
      journal={arXiv preprint arXiv:2511.21631},
      eprint={2511.21631},
      archivePrefix={arXiv},
      primaryClass={cs.CV} 
}

@inproceedings{zheng2024llamafactoryunifiedefficientfinetuning,
      title={{LlamaFactory}: Unified Efficient Fine-Tuning of 100+ Language Models}, 
      author={Yaowei Zheng and Richong Zhang and Junhao Zhang and Yanhan Ye and Zheyan Luo and Zhangchi Feng and Yongqiang Ma},
  booktitle={ACL},
  year={2024}
}

@inproceedings{sheng2024hybridflow,
  title   = {{HybridFlow}: A Flexible and Efficient RLHF Framework},
  author  = {Guangming Sheng and Chi Zhang and Zilingfeng Ye and Xibin Wu and Wang Zhang and Ru Zhang and Yanghua Peng and Haibin Lin and Chuan Wu},
  booktitle={EuroSys},
  year={2025}
}
